\documentclass[review]{elsarticle}

\usepackage{amsmath}
\usepackage{amssymb}
\usepackage{algorithm}
\usepackage{comment}
\usepackage{natbib}
\usepackage{mathtools}
\usepackage{graphicx}
\usepackage{tabularx}
\usepackage{adjustbox}
\usepackage{subcaption}
\usepackage{algorithmic}
\usepackage{varwidth}
\usepackage{xcolor}
\usepackage{booktabs}
\usepackage{pgfplotstable}

\usepackage{lineno,hyperref}
\modulolinenumbers[5]
\definecolor{RevCol}{RGB}{140, 20, 160}
\journal{peer review}

\makeatletter
\def\ps@pprintTitle{%
	\let\@oddhead\@empty
	\let\@evenhead\@empty
	\def\@oddfoot{}%
	\let\@evenfoot\@oddfoot}
\makeatother

\begin{document}
	
	\begin{frontmatter}
		
		\title{On-line conformalized neural networks ensembles for probabilistic forecasting of day-ahead electricity prices}
		
		\author[mymainaddress]{Alessandro Brusaferri\corref{mycorrespondingauthor}}
		\ead{alessandro.brusaferri@stiima.cnr.it}
		\author[mymainaddress]{Andrea Ballarino}
		\author[secondaddress]{Luigi Grossi}
            \author[thirdaddress]{Fabrizio Laurini}
		\cortext[mycorrespondingauthor]{Corresponding author}
		%\ead{alessandro.brusaferri@stiima.cnr.it}
		\address[mymainaddress]{CNR, Institute of Intelligent Industrial Technologies and Systems for Advanced Manufacturing, via A. Corti 12, Milan, Italy}
  	\address[secondaddress]{University of Parma, Department of Industrial Engineering, Viale delle Scienze, Parma, Italy}
    	\address[thirdaddress]{University of Parma, Department of Economics and Management, Via J.F. Kennedy 6, Parma, Italy}

		\begin{abstract}
		  Probabilistic electricity price forecasting (PEPF) is subject of increasing interest, following the demand for proper quantification of prediction uncertainty, to support the operation in complex power markets with increasing share of renewable generation. Distributional neural networks ensembles have been recently shown to outperform state of the art PEPF benchmarks. Still, they require critical reliability enhancements, as fail to pass the coverage tests at various steps on the prediction horizon.
          In this work, we propose a novel approach to PEPF, extending the state of the art neural networks ensembles based methods through conformal inference based techniques, deployed within an on-line recalibration procedure. 
          Experiments have been conducted on multiple market regions, achieving day-ahead forecasts with improved hourly coverage and stable probabilistic scores.
		\end{abstract}
		\begin{keyword}
		Probabilistic Forecasting \sep Electricity price \sep Day-ahead market \sep Neural Networks \sep Ensembles \sep Conformal Prediction \sep Time series
		\end{keyword}
		%\begin{keyword}
		%\texttt{elsarticle.cls}\sep \LaTeX\sep Elsevier \sep template
		%\MSC[2010] 00-01\sep  99-00
		%\end{keyword}
		
	\end{frontmatter}
	
	%\linenumbers
	
	\section{Introduction}
	\label{intro}
Trustworthy electricity price forecasting (EPF) systems represent fundamental strategic tools for utilities, retailers, aggregators, and large consumers to achieve effective participation in liberalized energy markets. Beside being exploited to anticipate price movements and perform bidding strategies, EPF is a key enabler of further crucial decision-making stages \cite{NOWOTARSKI20181548}, such as optimal generation and asset management \cite{MITRA2013194} and energy-aware planning and scheduling \cite{RAMIN2018622}. 
In particular, hourly day-ahead EPF - i.e., the prediction of the hourly prices for the next day - is a challenging task subject of continuous interest in both academia and industry \cite{LAGO2021116983}. Compared to other traded commodities, electricity is still not economically storable on a large scale; thus, a constant balance between demand and supply is essential to achieve overall system stability. Hence, the price profiles in liberalized power markets often exhibit quite peculiar characteristics \cite{CIARRETA2022}. Complex relationships with conditioning variables (such as load demand, fuel costs, electricity production and weather conditions) are typically involved, encompassing meshed short and long-term seasonalities, as well as considerable volatility, usually orders of magnitude larger than in other utility trading contexts \cite{LOUTFI2022119182}, \cite{WAGNER2022100246}, \cite{LEHNA2022105742}, \cite{9788043}.

Critical deviations have been injected by the repercussions of the steep gas price variations on the power plants. The increasing penetration of renewable sources in the generation mix, fundamental to contrast the global warming, is introducing further short-term price fluctuations to be promptly tackled. Additional variables such as the CO$_2$ certificate prices interact with the power price features \citep{MADADKHANI2024107241}. 
Additional complex market dynamics arise from the increasing competition among a growing number of highly active participants, each pursuing innovative business models \cite{ATTAR2022118905},\cite{TSCHORA2022118752}. 

Overall, forecasting is becoming harder to perform than ever before, and advances in this field are strongly requested to cope with such an evolving context \cite{LAGO2021116983}, \cite{KEZUNOVIC2020106788}.

Undoubtedly, the EPF research community is devoting considerable effort to addressing these key challenges, which is evident from the substantial amount of scientific studies developed in the literature, as detailed in the following subsection.

\subsection{Literature review}
Over the years, a broad set of approaches have been investigated to perform EPF. A non-exhaustive list includes auto-regressive models and related extensions, exponential smoothing, generalized additive models, Gaussian Process, gradient boosting, neural networks, support vector machines, random forest, fuzzy logic, wavelet ensembles, and hybrid models. More detailed descriptions and comparisons of these techniques can be found in \cite{WERON20141030}, \cite{LU2021100356},\cite{NOWOTARSKI20181548}, and references therein. 
According to the aforementioned reviews, the EPF literature was primarily dedicated to simple statistical models and small neural network architectures until the early 2010s, mainly due to computational constraints of the technologies available. Besides, they proved effective in addressing the more stable price series occurring during that period \cite{GIANFREDA20122228}. 

Following the demand for more enhanced modeling capabilities to properly tackle modern energy market dynamics, the EPF community has been investigated more flexible but computationally intensive machine learning approaches \cite{9218967}, \cite{TSCHORA2022118752}, \cite{LU2022118296}, \cite{FRAUNHOLZ2021116688}. 

Within this rapidly developing and fascinating field of research, much interest is being devoted to modern neural network architectures \cite{KELES2016218},\cite{GoodBengCour16}. Such research momentum is mainly propelled by the significant results achieved across a broad set of computer science applications, including natural language processing \cite{LAURIOLA2022443} and computer vision \cite{dosovitskiy2021an}.
Several research studies have contributed to the exploration and development of neural networks (NN) in the EPF field (see e.g., \cite{LAGO2021116983}, \cite{MASHLAKOV2021116405}, \cite{BRUSAFERRI20191158} for recent reviews).  
To date, the most comprehensive experimental investigation of NN based approaches to EPF is provided by \cite{LAGO2018386}, including feed-forward, recurrent and convolutional models, as well as several more traditional statistical techniques (such as ARX, ARIMA-GARCH, etc.), primarily adopted in the literature. When tested on the EPEX (Belgium day-ahead market benchmark), a deep feed-forward architecture (labeled DNN hereafter) achieved the highest predictive accuracy, outperforming both the traditional methods and the best recurrent neural networks with LSTM and GRU cells \footnote{LSTM: Long short-term memory, GRU: Gated recurrent unit}.

Building upon the previous works, in \cite{LAGO2021116983} and \cite{9761111} the authors have conducted a comparison betwee the DNN  architecture and the recently proposed LASSO Estimated AutoRegressive model, which is arguably the most accurate approach within the broader statistical family. By deploying an extensive automatic hyperparameter optimization and evaluating a set of benchmark EPF tasks covering five different day-ahead markets, they confirmed the performance gains provided by the more flexible DNN. Indeed, the latter paper has been entitled “the dawn of machine learning” for EPF.  
Following these results, several NN-based techniques have been explored in recent years with the aim of reducing prediction error (see e.g., \cite{OLIVARES2023884} and references therein).
In economic terms, even single percentage gains can lead to annual savings of up to millions of dollars for companies \cite{KEZUNOVIC2020106788}.

Although most of the available studies address point predictions (e.g., mean or median values) of day-ahead hourly prices, an increasing research momentum targets NN-based probabilistic EPF (PEPF). Specifically, PEPF is aimed to characterize the underlying uncertainty of the model in the target space, e.g., via prediction intervals, quantiles or full predictive distributions. 
This is essential for companies operating within power markets showing increasing volatility, e.g., to support extremely relevant business tasks, such as effective risk optimization, stochastic optimization and what-if scenario analysis before trading. Several recent studies investigate this topic from different perspectives, reporting both additional problems and opportunities insisting on the EPF approach (see e.g., \cite{DUMAS2022117871}, \cite{GROTHE2023106602}, \cite{10012043}, \cite{CRAMER2023121370} and references therein for details). 
An thorough empirical comparison of deep learning architectures for multivariate probabilistic energy forecasting is performed in \cite{MASHLAKOV2021116405}, including DeepAR, DeepTCN, DSANet and LSTNet models.

According to several research works (see e.g., \cite{NOWOTARSKI20181548},\cite{MARCJASZ2023106843}, \cite{KATH2021777} and references therein), Quantile Regression Average (QRA) represents the reference benchmark to turn point NNs into a PEPF setup, following the ranking in the GEFCom competition \footnote{The Global Energy Forecasting Competition (GEFCom) began in August 2014, with a specific emphasis on probabilistic energy forecasting, including load, price, wind, and solar predictions. The price category drew interest from 287 participants globally, with the top-performing teams subsequently invited to contribute papers to a special 2016 edition of the International Journal of Forecasting \cite{HONG2016896}}. 

Distributional NNs \cite{MASHLAKOV2021116405} and Bayesian Deep learning \cite{BRUSAFERRI20191158} techniques have been proposed approximating the conditional distribution of the hourly prices given the input features, through parametric forms. The Gaussian distribution is commonly employed for this purpose. However, it has been observed that the prices tend to exhibit complex patterns, including heterosckedasticity, significant skewness and fat tails \cite{MARCJASZ2023106843} and \cite{9263898}. Deep quantile regression (DQR) based methods have been introduced to support non-parametric uncertainty characterization (see e.g., \cite{ZHOU2022101489} and references therein). Despite their asymptotic properties, they inherit potential quantile overfitting and overconfidence on the extreme quantiles in finite samples. Hence, Ensembles of distributional NNs (i.e., Deep Ensembles) parameterizing flexible Johnson’s SU distributions have been recently proposed for this purpose, outperforming the state of the art PEPF benchmarks including both conventional Gaussian forms and the widely applied quantile regression on NN ensembles \citep{MARCJASZ2023106843}. 
Despite the relevant improvements, it has been observed that such models lack the required calibration capabilities - i.e., the statistical consistency between the probabilistic forecasts and the observations - failing to pass the coverage tests at various hours on the prediction horizon.

Conformal Prediction (CP) provides a principled framework to attain distribution-free finite sample marginal calibration guarantees. Pioneered in the early 2000s, CP has recently become subject of a renewed research interest within the statistical and machine learning communities (see e.g., \cite{b2} and reference therein for a detailed review). However, CP has so far attracted minor attention in the EPF field. To date, only two papers have investigated CP-based approaches for this purpose. A first empirical study has been performed by \cite{KATH2021777} where the conventional split CP is compared to a
normalized absolute deviation on LASSO, KNN, SVM models, achieving promising results. Later, \cite{pmlr-v162-zaffran22a} implemented an adaptive conformal inference procedure on Random Forest based point predictors.
As this presents a relatively novel and still understudied research direction, further extensions are expected from the community in the near future.

By looking beyond the EPF literature, last developments in the Conformal Prediction research domain provides a lot of compelling ideas to be further explored for achieving sharp and calibrated probabilistic forecasts (see e.g., \cite{angelopoulos2022gentle} and references therein).
To the best of our knowledge, the state-of-the-art neural network-based approaches for PEPF have not yet been compared with CP extensions to assess potential probabilistic performance gains.

\subsection{Contributions and organization of the paper}
Leveraging the available state of the art and moving from reported open issues, the major scope of the present work is to contribute to the further investigation of the CP framework within the PEPF field. 

Specifically, we target the approximation of the feature conditioned day-ahead prices distribution through a discrete set of quantiles (e.g., deciles), coupled in prediction intervals (PIs) of increasing coverage degree around the median. 

Then, we propose a novel approach to PEPF, extending the Deep Ensembles based methods through conformal inference based techniques deployed within an on-line recalibration procedure. 
The developed method consists of the following major ingredients.
First of all, we leverage the asymmetric CP formulation introduced by \citep{NEURIPS2019_5103c358} for regression tasks, here developed through a daily recalibrated multi-horizon time series setup to support flexible step-wise compensations of each upper/lower band. Overall, this enhance local PIs reliability (i.e., sample-wise efficiency) beyond the CP marginal coverage.
Beside, we deploy the adversarial conformal inference setup proposed in \cite{angelopoulos2023conformal} for uncertainty quantification under non-exchangeable conditions, such as distribution shifts. Hence, the target prediction bands over the different coverage levels are dynamically adjusted by tracking the related quantiles of the conformity score sequences, incorporating the running sum of the miscoverage events to support a stable long-run calibration.
To estimate the quantiles to be calibrated, we explore both a DQR setup and samples from different parameterized distributional NNs to assess their potential impact on both sample efficiency and reliability. A uniform vincentization technique is exploited for ensembles aggregation, leading to sharper bounds than the alternative probability aggregation, beside marginalizing the different local minimizers reached by the training algorithm. Moreover, a post-hoc sorting operator is included before combination to achieve conditional quantile non-crossing.

We structured an open GitHub repository with the datasets and code to reproduce the experiments, as well as to support the integration and comparison of further datasets and PEPF techniques \footnote{https://github.com/bruale/PefCodeBench}. 

The experiments are performed following the established best practice guidelines stated in \cite{LAGO2021116983} for EPF research. 
As case studies, we focused on the German market dataset (i.e., the benchmark application in the original Distributional NN paper \cite{MARCJASZ2023106843}), as well as the different regional bidding zones constituting the Italian day-ahead markets made available by \cite{forecast5010003}, providing a compelling setup for testing under heterogeneous conditions. 
A comparison against state of the art benchmarks is performed, including QRA, DQR, Distributional NNs, as well as conventional absolute score and normalized score based CP settings.

The rest of the paper is structured as follows. Section II deepens the developed PEPF approach, while articulating the techniques in mathematical form. Section III provides a detailed description of the experimental setups and reports the results achieved. Section IV summarizes the conclusions and the envisioned future works.

	\section{Methods}
	\label{Methods}
In this section, we deepen the developed PEPF approach. We start by providing a brief introduction to the general conformal inference framework. 
Then, we address the local adaptivity issue by deepening the conformalized quantile regression technique. Subsequently, we describe the methods deployed to address the lack of robustness of CP under non-exchangeable conditions.
Finally, we detail the prediction quantiles estimation procedures, the network archiectures employed, as well the deep ensemble combination techniques.

\subsection{Conformal Prediction}
\begin{figure}[b!]
    \centering
    \includegraphics[width=1\linewidth]{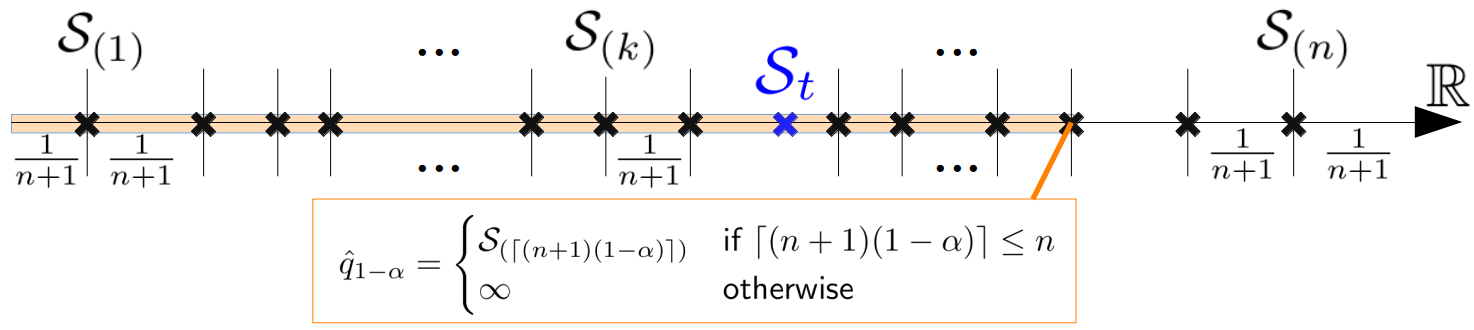}
    \caption{Empirical quantiles of the conformity scores}
\label{fig1}
\end{figure}
Conformal Prediction (CP) provides a general framework to obtain prediction intervals $\mathcal{C}(\mathcal{D}_n, \alpha, x_{t}) \equiv \mathcal{C}(x_{t})$, at any test point $t$, from black-box models with marginal coverage guarantees under finite sample settings. Formally, this is stated as:
$\mathbb{P}(y_{t} \in \mathcal{C}(x_{t})) \geq 1-\alpha, \text{ with $\alpha$-error probability}$,
given a dataset $\mathcal{D}_n \equiv \{(x_i, y_i)\}_{i=1}^n$, $y_i \in \mathbb{R}$ response variable and features vector $x_i=[x_i(1),...,x_i(d)]$ involving continuous or discrete components.
Moreover, CP is fully distribution-free, as opposed to distributional neural networks techniques (parameterizing e.g. Gaussian, Student's t, mixtures, etc.), thus retaining validity for arbitrary latent distributions. 
Clearly, this is trivial without efficiency requirement (e.g., $\mathcal{C}(x_{t})=\mathbb{R} \Rightarrow \mathbb{P}(.)=1$). Hence, the goal is to achieve a sharp interval closed to the equality $\mathbb{P}(y_{t} \in \mathcal{C}(x_{t}))\approx 1-\alpha$.

The core concepts behind CP are the conformity scores, which are exploited to assess the degree of "conformity" (thus the name of the approach) of the trained model prediction with reference to an held-out calibration bag (i.e., split CP). For continuous target values, the absolute score $\mathcal{S}(x_i,y_i)=|y_i - f(x_i)|$ is conventionally employed, where $f(x_i)$ represents the model sample prediction. 

Then, by computing the empirical quantiles of the order statistic obtained by ranking the conformity scores (depicted in Figure \ref{fig1}), it is easy to show that for any $t$-th test sample:
\begin{align}
    &\mathbb{P}(|y_t - f(x_t)| \leq \mathcal{S}_{(\lceil(n+1)(1-\alpha) \rceil)})= \nonumber \\
    &\mathbb{P}(y_t \in \underbrace{f(x_t) \pm \mathcal{S}_{(\lceil(n+1)(1-\alpha) \rceil)}}_{C_{1-\alpha}(x_t)})=\frac{\lceil(n+1)(1-\alpha) \rceil}{n}
    \label{eq1}
\end{align}
Besides, using a random tie-breaking rule (to avoid ties in the absolute residuals ranking), it has been shown that:
$\mathbb{P}(y_{t} \in \mathcal{C}_{1-\alpha}(x_{t})) \leq 1-\alpha + 1/(n+1)
$. Further details on CP and related proofs are reported in \cite{b2} and references therein.

\subsection{Improved local adaptivity by conformalized quantiles}
\label{local_adapt}
As apparent from the Eq.\ref{eq1}, the conventional CP based on absolute scores results in symmetric average bands around the NN predictions, thus lacking the capability to adapt the PIs width on simpler/harder test conditions (e.g., heteroscedasticity). Since full conditional coverage (i.e., $\mathbb{P}(y_{t} \in C(x_{t})|x_{t}=\chi) \geq 1-\alpha$, for almost all $\chi$) is not reachable in finite samples under weak distribution-free assumption, the goal is to obtain sample-wise efficiency (local PIs adaptivity) with proven marginal validity. 

To this end, two major classes of methods have been recently proposed in the CP literature, namely normalized conformal inference-based, and conformalized quantile regression (CQR)-based approaches (see e.g., \cite{angelopoulos2022gentle} and references therein).  
The former typically rescales the CP scores depending on the conditional spread of the observations, e.g., by paramaterizing the target variance via dedicated network outputs as in distributional NNs. Still, this procedure leads to symmetric bounds around the point predictions despite the underlying distribution shape.
Furthermore, it has been observed to suffer from critical PI inflation and systematic underestimation issues \cite{angelopoulos2022gentle}.

In consideration of these results in the literature, we deploy a CQR-based approach in the PEPF framework.
Specifically, to compute the scores and the conformalized quantiles, we leverage the asymmetric form introduced by \cite{NEURIPS2019_5103c358}, which can be easily adapted to the multi-horizon forecasting setup proposed in \cite{NEURIPS2021_312f1ba2} as follows:
\begin{equation}
  \mathcal{C}^h_{1-\alpha}(x_{t})=[{q}^h_{\alpha/2}(x_t)-{l}^h_{1-\alpha/2}(\mathcal{I}_{c_t}), \ \ {q}^h_{1-\alpha/2}(x_t)+{u}^h_{1-\alpha/2}(\mathcal{I}_{c_t})] 
\end{equation}
where $\mathcal{C}^h_{1-\alpha}(x_{t})$ represents (1-$\alpha$)-level PI at prediction step $h=[1,...,H]$ given the input features. 
$\small{{l}^{h}_{1-\alpha/2}\small}$ and ${u}^h_{1-\alpha/2}$ depict the (1-$\alpha/2$)-th empirical quantiles of ${q}^h_{\alpha/2}(x_i)-y_i^h: i \in \mathcal{I}_{c_t}$ and $y_i^h-{q}^h_{1-\alpha/2}(x_i): i \in \mathcal{I}_{c_t}$ respectively, computed on the calibration subset $\mathcal{I}_{c_t}$ for test time $t$. 
$\{{q}^{h}_{{\gamma}}(x_j)\}_{{\gamma} \in \Gamma}, \: {q}^{h}_{{\gamma}}(x_j) \leq {q}^{h}_{{\gamma}'}(x_j) \; \forall {\gamma} <{\gamma}'$ represent the discrete set $\Gamma$ of the quantiles predicted by the NN ensemble (see section \ref{DE comb}), while  $y_t^{{h}}$ are the observed values.
The PI bounds at specific coverage levels are straight derived as the predicted quantiles in the set $\Gamma$ are defined by balanced pairs (e.g., distribution deciles, percentiles, etc).
   % $\mathcal{C}^h_{1-\alpha}(x_{i})=[\tilde{Q}^h_{\alpha/2}(x_i)-\hat{q}_{1-\alpha/2}(\mathcal{S}^h_{lo},\mathcal{I}_c),\tilde{Q}^h_{1-\alpha/2}(x_i)+\hat{q}_{1-\alpha/2}(\mathcal{S}^h_{up},\mathcal{I}_c)]$\\[4pt]
   % $\hat{q}_{1-\alpha/2}(\mathcal{S}^h_{lo},\mathcal{I}_c):1-\alpha/2$-th empirical quantile of $\{\tilde{Q}^h_{\alpha/2}(x_i)-y_i^h: i \in \mathcal{I}_c \}$\\[2pt]
   % $\hat{q}_{1-\alpha/2}(\mathcal{S}^h_{up},\mathcal{I}_c):1-\alpha/2$-th empirical quantile of $\{y_i^h-\tilde{Q}^h_{1-\alpha/2}(x_i): i \in \mathcal{I}_c \}$\\[6pt]

Thanks to such formulation, upper/lower quantiles can be specifically adjusted at each stage over the prediction horizon, considering the related coverage on the calibration sets. Overall, this yields stronger coverage capabilities by enabling more efficient input features conditioned approximate PIs estimation beyond the CP marginal coverage. In fact, more flexibility in compensating for the predicted bands is deemed useful for addressing the complex shapes typically involved in PEPF applications, such as heteroscedasticity, skewness and fat tails \cite{MARCJASZ2023106843}.
Besides, the asymptotic validity supported by quantile regression NNs (i.e., DQR) is extended to reach finite samples coverage guarantees, while compensating for potential quantile overfitting and overconfidence.

\subsection{Addressing non-exchangeable conditions}
\label{non-exc}
Still, the CQR-based formulation inherits the lack of robustness of CP to failures of the samples exchangeability assumptions. This simply states that the target (arbitrary) joint distribution is invariant to observations permutation, thus representing a weaker assumption than the i.i.d. commonly employed in machine learning applications (see e.g., \cite{CP_unified_rev} for further details). 
In particular, covariates and concept shifts in the target series are recognized as critical issues impacting CP validity in practical time series applications, as it is the case for EPF.

Three main classes of approaches have been recently investigated in the CP literature devoted to non-exchangeable time series tasks, namely online sequential split CP (see e.g. \cite{KATH2021777}), bootstrapped estimators without underlying models refitting (as e.g., in \cite{EnbPI},\cite{9940232}) and adversarial CP settings (see e.g., \cite{NEURIPS2021_ACI},\cite{pmlr-v162-zaffran22a},\cite{angelopoulos2023conformal}). 

In this work, we integrate the multi-step CQR-based compensations (reported in section \ref{local_adapt}) within an on-line NN ensemble (i.e., Deep Ensemble) recalibration procedure, including the quantiles tracking and coverage error integration stages proposed in \cite{angelopoulos2023conformal} (i.e., Conformal PI control). 
Specifically, the quantiles predicted by the Deep Ensemble (DE) are incrementally conformalized over the test set following a daily retraining (i.e., recalibration). Details about how prediction models recalibration is commonly employed in the EPF context can be found in \cite{LAGO2021116983}.
Formally, the predicted quantiles are corrected as follows at each stage $t$ (i.e., day) in the test set:
\begin{align}
    \bar{q}^h_{t+1}&=\bar{q}^h_{t}+\eta \nabla \rho_{1-\alpha/2} \left(s^h_t-\bar{q}^h_{t}\right) + r_t\left[\sum_{j=1}^t (1\{\bar{q}^h_{j}\geq s^h_j\}-\alpha/2)\right]\\
    r_t(x)&=K_I\text{tan}\left(\frac{x\text{log}(t)}{tC_{sat}}\right)
\end{align}
where $s^h_t$ summarize the asymmetric CQR scores, $1\{.\}$ the indicator function, and $\nabla \rho_{1-\alpha/2}(.)$ denotes the subgradient of the quantile loss, with:
\begin{equation}
    \nabla \rho_{1-\alpha/2}(s^h_t-\bar{q}^h_{t})=
    \begin{cases}
        1-\alpha/2 &\text{if: } s_t^h>\bar{q}_t^h \\
        -\alpha/2 &\text{otherwise}
    \end{cases} 
\end{equation}
The operation is performed on both the (1-$\alpha/2$)-th empirical quantiles of the asymmetric upper/lower scores ${q}^h_{\alpha/2}(x_t)-y_t^h$ and $y_t^h-{q}^h_{1-\alpha/2}(x_t)$.
We keep $r_t(x)$ as the tangent integrator, leaving the investigation of alternative admissible saturation functions to future extensions. 
The constant $C_{sat}$ bounds the asymptotic guarantee of the integral action at a level of at least $1 - \alpha - \delta$:
\begin{equation}
    C_{sat}=2/\pi\left[ \lceil\text{log}(T)\delta\rceil-1/\text{log}(T)\right]
\end{equation}
with $\delta>0$ representing a small constant (e.g., 1e-2).
$\eta$, $K_I>0$ are further tunable hyperparameters controlling the proportional and integral actions respectively(see section \ref{Results}).

The whole procedure is first executed across a sub-sequence of samples close to the first test date to acquire the initial bag of calibration scores. It then progresses in a rolling window fashion.

The motivations behind such methodological choices are summarized hereafter. First of all, while in principle CP yields marginal coverage despite the accuracy of the underlying model, the latter impacts the local reliability of the prediction intervals \cite{angelopoulos2022gentle}. 
Hence, updating the DE by exploiting the last observations can support probabilistic performances beside point accuracy.
Besides, the ensemble combination (detailed in section \ref{DE comb}) provides a simple but efficient technique to get rid of poor local minimizers in the network parameter space (see section), often leading to improved accuracy beyond single networks (see e.g., \cite{LAGO2021116983}). 
Despite more computationally demanding, online DE retraining is feasible for EPF applications (see e.g., \cite{MARCJASZ2023106843}). For instance, it is commonly employed within the widely adopted Quantile Regression Averaging (QRA) technique, as well as for distributional NNs aggregations in the litarature.
The investigation of conformalized quantile regression methods without DE recalibration (as e.g. in \cite{9940232}) and further on-line adaptive CP wrappers (as e.g., \cite{pmlr-v162-zaffran22a}) is left to future works.
Finally, the quantiles tracking and integration provides further mechanism to compensate coverage degradation and systematic errors occurring over the recalibration window, e.g., due to sensible short-term drifts. This support a principled long-run reliability, i.e.,
$\lim_{T\to +\infty}1/T \sum_{t=1}^T 1\{y_t^h \not\in \mathcal{C}^h_{1-\alpha}(x_{t})\}=\alpha$, for each target coverage degree $1-\alpha$ (see \cite{angelopoulos2023conformal} for further details on long-run coverage and related proofs). The exploration of the derivative action (i.e., the scorecasting component) is left to future works devoted to specific robust design for addressing the potential degrade in stability. 

\subsection{Prediction quantiles estimation and DE combination}
\label{DE comb}
To first estimate the conditional quantiles to be calibrated, we exploit both a quantile regression setup and samples from distributional NNs. In principle, the former enables flexible non-parametric approximations, but suffers potential quantiles overfitting and overconfidence in the extremes under finite date regimes \cite{NOWOTARSKI20181548}. The latter can mitigate such issues by leveraging the parametric form,  but requires a careful selection of the density for each task at hand. Therefore, a dedicated experimental comparison is deemed useful to assess their capabilities in practical conditions. 

The output layers of the NNs and the loss functions are specialized accordingly, i.e. providing the predicted quantiles averaged by the pinball score in the former, parameterizing the target distribution then passed through a negative log-likelihood in the latter. 

The next step in designing the PEPF framework regards the specification of the NN architectural form employed to compute the day-ahead predictions.
Either single-step (i.e., using 24 hour specific models) and multi-step mapping forms can be considered for this purpose (see e.g., \cite{LAGO2018386}). Besides, as introduced in section \ref{intro} a broad amount of architectures have been investigated in the EPF literature, including feed-forward, recurrent NNs and Transformers to cite a few (see e.g., \cite{MASHLAKOV2021116405} and references therein for further details). 

As the state-of-art Distributional NN for PEPF is based on multi-step feed-forward maps, we follow the same structural design in this work. 
This enables a proper comparison of the proposed PEPF technique under coherent settings (see section \ref{Results}). The investigation of further NN architectural forms is left to future studies.
Such NN configuration is labelled DNN hereafter (i.e., Deep Neural Network), following the naming convention in \cite{MARCJASZ2023106843} \cite{LAGO2021116983})

Formally, each DE component is defined as a parameterized function providing day-ahead price predictions over the whole horizon in a unique pass,
given the input conditioning set $x_{i}$ at time $i$. 
Considering two hidden layers of $n_{u_1}, n_{u_2}\in \mathbb{Z}^+$ units to lighten the notation, the
feed-forward map is mathematically expressed as:
\begin{align}
\label{DNN_eq}
	\begin{aligned}
        \ell_1 &=\text{g}(x_i W_1 + b_1) \\
        \ell_2 &=\text{g}(\ell_1 W_2 + b_2)W_3 + b_3 \\
	\end{aligned}
\end{align}
where $W_1 \in \mathbb{R}^{n_x \times n_{u_1}}, \ W_2 \in \mathbb{R}^{n_{u_1} \times n_{u_2}}, \ W_3 \in \mathbb{R}^{n_{u_2} \times H\cdot n_p},b_1\in \mathbb{R}^{n_{u_1}}, b_2\in \mathbb{R}^{n_{u_2}}, b_3\in \mathbb{R}^{H\cdot n_p}$ represent the weight and bias parameters in each hidden layer and $g(.)$ the nonlinear activation function (e.g., ReLU, LeakyReLU, GELU, etc.). 

The input feature $x_i$ can involve both past values of the target series and a set of exogenous variables (see \ref{Results} for the sets involved in each benchmark application). Hence, the learning machinery is aimed at identifying the latent relationships and patterns within the spreading information sources involved in the conditioning set \cite{LAGO2018386}. 

The parameter $n_{p}\in \mathbb{Z}^+$ is defined depending on the subsequent network setup. Specifically, it is set as:
\begin{itemize}
    \item $n_{p}=1$ : to achieve point predictions at each hour, as employed e.g., in the QRA-based PEPF setup (see section \ref{Results});
    \item $n_{p}=\#\Gamma$ : for quantile regression settings, with $\#\Gamma$ representing the number of estimated quantiles (e.g., $\#\Gamma=10$ for deciles approximation);
    \item $n_{p}=p$ : for distributional NN settings, where $p$ represent the number of parameters of the approximated conditional distribution, e.g., $p=4$ for the Johnson’s SU form (see section \ref{Results}). 
\end{itemize}

For the Distributional NN configurations, the output of the last layer in \ref{DNN_eq} (i.e., $\ell_2$) are further processed before being passed to the succeeding distributional layer in order to achieve valid parameterizations. For instance, in the case of the Johnson’s SU form:
\begin{equation}
    f^h(\chi)= \frac{\tau_i^h}{\sigma_i^h\sqrt{2\pi}}\frac{1}{\sqrt{1+\left(\frac{\chi-\lambda_i^h}{\sigma_i^h} \right)^2}}e^{-\frac{1}{2}\left[ \zeta_i^h + \tau_i^h \text{sinh}^{-1}\left(\frac{\chi - \lambda_i^h}{\sigma_i^h} \right) \right]^2}
\end{equation}
the output of the last hidden layer is processed as:
\begin{align}
\label{jsu_tr}
    &\lambda_i^h = \ell_2^{[h]}\\
    &\sigma_i^h = \epsilon + \gamma\text{ Softplus}\left(\ell_2^{[H+h]}\right)\\
    &\tau_i^h = 1 + \gamma\text{ Softplus}\left(\ell_2^{[2\cdot H+h]}\right)\\
    &\zeta_i^h = \ell_2^{[3\cdot H+h]}\\
    &\text{Softplus}(x)=\text{log}\left(1+e^{x} \right)
\end{align}
with $\lambda_i^h,\sigma_i^h,\tau_i^h,\zeta_i^h$ defining the density location, scale, tailweight and skewness given by the feed-forward map, conditioned on the input feature $x_i$. $\ell_2^{[j]}$ defines the element in the output vector $\ell_2$ at index $j$. $\epsilon=1e^{-3}$ and $\gamma=3$ are correction factors commonly introduced for computational purpose.
As apparent by the formulations above, a specific distribution parameterization is computed at each stage over the prediction horizon.
Similar transformations are employed for the other density forms, e.g., to constrain positive values for standard deviation and degrees of freedom in Normal and Student-t (see section \ref{Results}).

For the quantile regression setup, the predicted conditional quantiles (e.g., deciles) at each stage $\hat{q}^{{h}}_{\gamma}(x_i)$ are simply extracted by indexing the last layer $\ell_2$ in the feed-forward map by stage $h$ and level $\gamma$. 
Then, the average Pinball loss across the discrete set $\gamma \in \Gamma$ of target quantiles is computed as:
\begin{equation}
\sum_{i}{\sum_{h}}\sum_{\gamma}(y_i^{{h}}-\hat{q}^{{h}}_{\gamma}(x_i))\gamma 1\{y_i^{{h}}>\hat{q}^{{h}}_{\gamma}(x_i)\} + (\hat{q}^{{h}}_{\gamma}(x_i)-y_i^{{h}})(1-\gamma) 1 \{y_t^{{h}}\leq \hat{q}^{{h}}_{\gamma}(x_i)\}
\end{equation}

Still, the approximated conditional quantiles can be subject of critical crossing issues. To address this problem, the introduction of non-crossing constraints within the quantile loss function has been proposed in the literature. Despite providing a valuable solution, it results in a sensible increase of the computational effort during learning. Besides, it does not provide non-crossing guarantees for the whole conditioning space. Therefore, in this work we exploit a post-hoc sorting operator, formally defined as:  
\begin{equation}
    \{\tilde{q}^{h}_{{\gamma}}(x_i)\}_{{\gamma} \in \Gamma}=\text{Sort}(\{\hat{q}^{h}_{\gamma}(x_i)\}_{\gamma \in \Gamma}), \: \tilde{q}^{h}_{{\gamma}}(x_i) \leq \tilde{q}^{h}_{{\gamma}'}(x_i) \; \forall {\gamma} <{\gamma}'
\end{equation}
Beside providing a computationally cheap solution to achieve non-crossing $\forall x$ during prediction by construction, it has been recently shown that such operation can only improve the pinball loss in the post-processed forecasting model \cite{fakoor2023flexible}.

The last design stage regards the definition of the approach to combine the neural network components within the DE. To this end, a wide set of methods have been studied in the time-series forecasting literature, from uniform marginal aggregations, to complex conditional weighting (see \cite{WANG20231518} for a recent detailed review). Still, the uniform aggregation (i.e., ensemble combination considering equal weights) have been shown to provide a robust alternative difficult to beat in practical settings. Similar observations have been reported under both point \cite{LAGO2021116983} and probabilistic \cite{MARCJASZ2023106843} DNN-based settings EPF studies. Moreover, it constitute a widely adopted approach in the broader Deep Learning literature devoted to DE (see e.g., \cite{NIPS2017_9ef2ed4b})
Therefore, we exploit a uniform ensemble combination for the present study, leaving the investigation of alternative techniques to future extensions.
For point predictors (e.g., in the case of QRA and base CP in the next section), this results in a simple average between the DNN components.
For Distributional NNs, both probability aggregation and vincentization techniques can be considered. We resort to the latter since it typically leads to sharper bounds, beside marginalizing the different local minimizers reached by the training algorithm \cite{WANG20231518}. 
Formally, this is computed for each prediction stage and quantile as:
\begin{equation}
    q^h_\gamma(x_i)=\sum_{j=1}^{n_e} \tilde{q}_{{\gamma}}^{h\mathbf{(j)}}(x_i), \; \forall {\gamma} \in \Gamma, \; \forall {h} \in [1,...,H]
\end{equation}
where $n_e$ represents the number of neural networks constituting the ensemble, indexed by $(j)$.
The same approach is employed for forecasting combination in the quantile regression configuration.

The overall DE components are trained end-to-end in parallel, starting from different random initializations, by passing evenly spaced batches of data including the whole set of conditioning features and the target price values. To form the batches from the input time series during daily recalibration, we employ a common sliding window approach, as detailed in Section \ref{Results}.

	\section{Applications and results}
	\label{Results}
\subsection{Case studies}
As case studies, we focused on both the German market (GE) from \cite{MARCJASZ2023106843} and the different bidding zones constituting the Italian day-ahead markets (namely NOR, CNOR, CSOU, SOU, SARD, SICI) made available by \cite{forecast5010003}, providing a compelling setup for comparative evaluations under heterogeneous conditions. The exogenous set for the GE market includes the day-ahead load and renewable generation forecast as well as the most recent gas closing prices. The Italian datasets comprise the hourly load and wind generation predictions. 

The German dataset spans observations from 1/1/2015 to 31/12/2020, with out-of-sample test starting on 27/6/2019. The last 364 days before the beginning of the recalibration warm-up (which starts on 27/12/2018) are employed as validation data for hyper-parameters tuning.
%\textbf{In this section, we aim to adhere as closely as possible to the framework used in the two cited papers for two primary reasons: 1) ensuring that all results are fully replicable, and 2) enabling direct comparison of the forecasting performance of the methods employed in this paper with the results obtained in the cited papers, from which the data are sourced.}

As input features, we included the subset selected with frequency 100\% during the experiments performed in \cite{MARCJASZ2023106843} 
To summarize, we adopted: the price values over the previous 2 days $t$-1, $t$-2, i.e., 48 lags; the day-ahead load forecasts available at time $t$; the renewable energy sources predictions for the target date $t$ as well as the previous day $t$-1; the most recent closing gas price available, i.e. at $t$-2; the weekday encoding.
The motivation behind such choice is twofold. On the one hand, we aim to explore a group similar to the one employed in the previous work. On the other hand, we target the assessment of the different PEPF models under consistent input variables.\footnote{{In \cite{MARCJASZ2023106843}, the authors reported sensible variations both in terms of the hyper-parameters chosen (including NNs input features selection via indicator variables) and the consequent test performances (see \cite{MARCJASZ2023106843} for further details). Still, a deepen dataset analysis and feature selection may lead to better average performances. We leave such investigation to future extensions.}}

The Italian datasets cover the period from 10/1/2015 to 31/8/2019. For each region, the test set starts from 31/8/2018 onwards, while the validation subset covers the last year of observations before the first test recalibration. Considering the analysis reported in \cite{forecast5010003}, we have included in the input features the price settlements across the different hours of the last 7 days (hence 168 values), each of the 24 values for both day-ahead load and wind generation forecasts, as well as the weekday.

The weekday input is represented by means of the cyclical feature encoding in sine-cosine form, computing the components in the vector:
\begin{equation}
    c_{t}=  \left[ \text{sin}\left(2\pi d_t/7 
  \right), \ \ \text{cos}\left(2\pi d_t/7 
  \right) \right]
\end{equation}
where $d_t\in[0,...,6]$ indexes the day of the week for the predicted sample at time $t$.\footnote{The investigation of alternative techniques (e.g., one-hot encoding) is left to future studies.}
The targets include the price values for each of the day-ahead 24 hours at stage $t$.

The samples are generated from the time series through a moving window. A Z-Score normalization is applied within each recalibration run, fitted by involving only the past values. Besides, a batch normalization is inserted after the DNN input layer. 

\subsection{Benchmarks}
The proposed approach is compared to several state of the art DNN based PEPF methods, including QRA, Deep Quantile Regression (QR) and Distributional NNs. In addition to the Johnson’s SU (Jsu) proposed in \cite{MARCJASZ2023106843}, we deploy a Student's $t$ (Stu) to investigate the impact of the different distributional setups on the case studies.

Moreover, we implement CP techniques on both normal distributions and conventional absolute residuals (see \cite{KATH2021777}) to assess the benefit of the more flexible quantile-level corrections introduced. 
The former has been implemented following the same setup of the other distributional NN models to achieve a fair comparison. Hence, each NN in the ensemble parameterizes the mean and scale of the Gaussian distribution, followed by quantile vincentization. 
Although normalized conformity scores may be alternatively computed through the mean and variance of a mixture with uniform weights, probabilistic aggregation has been shown to perform worse than quantile ensembling (see \cite{MARCJASZ2023106843}, \cite{Lichtendahl}). \footnote{{The exploration of conformal inference on further forecasting combination techniques represent an interesting direction of future research.}}
The CP on absolute residuals is obtained from the average predictions of NNs trained by minimizing the Mean Absolute Error.  

To shorten notation, in the subsequent sections we employ the following labels for referring to the different PEPF methods: CQN for the parameterized Normal form (i.e.,Conformalized Quantile Normal); CQR for Conformalized Quantile Regression; CQJ for Conformalized Quantile Johnson’s SU; CQS for Conformalized Quantile Student's t. The application of the quantiles tracking and coverage error integration stages is represented by stacking the label OCQ*, e.g., OCQJ for on-line conformal PI control on Jsu distributional NNs samples quantiles.    

\subsection{Experimental setup}
\label{exper_sub}
\begin{table}[t!]
\caption{Hyperparameters selected by the grid-search procedure for each zone}
\label{hyper_tune}
\begin{center}
\begin{adjustbox}{width=1\textwidth}
\small
\begin{tabular}{lllllllllll}
\bf{GE} &CP/QRA &Norm/CQN/OCQN &Jsu/CQJ/OCQJ &Stu/CQS/OCQS &QR/CQR/OCQR 
\\ \hline
${n_h}$ &768 &896 &896 &640 &640 \\ 
${l_r}$ &1e-3 &1e-3 &1e-3 &1e-3 &1e-4 \\ \hline
\bf{NOR} &CP/QRA &Norm/CQN/OCQN &Jsu/CQJ/OCQJ &Stu/CQS/OCQS &QR/CQR/OCQR
\\ \hline
${n_h}$ &512 &768 &640 &640 &896 \\ 
${l_r}$ &1e-3 &1e-3 &1e-3 &1e-3 &1e-4 \\ \hline
\bf{CNOR} &CP/QRA &Norm/CQN/OCQN &Jsu/CQJ/OCQJ &Stu/CQS/OCQS &QR/CQR/OCQR
\\ \hline
${n_h}$ &768 &960 &896 &896 &896 \\ 
${l_r}$ &1e-3 &1e-3 &1e-4 &1e-3 &1e-4 \\ \hline
\bf{CSOU} &CP/QRA &Norm/CQN/OCQN &Jsu/CQJ/OCQJ &Stu/CQS/OCQS &QR/CQR/OCQR
\\ \hline
${n_h}$ &640 &960 & 896 &768 &960 \\ 
${l_r}$ &1e-3 &1e-3 &1e-4 &1e-3 &1e-4 \\ \hline
\bf{SOU} &CP/QRA &Norm/CQN/OCQN &Jsu/CQJ/OCQJ &Stu/CQS/OCQS &QR/CQR/OCQR
\\ \hline
${n_h}$ &768 &896 &896 &960 &960 \\ 
${l_r}$ &1e-4 &1e-3 &1e-4 &1e-4 &1e-4 \\ \hline
\bf{SARD} &CP/QRA &Norm/CQN/OCQN &Jsu/CQJ/OCQJ &Stu/CQS/OCQS &QR/CQR/OCQR
\\ \hline
${n_h}$ &896 &960 &960 &640 &896 \\
${l_r}$ &1e-3 &1e-3 &1e-3 &1e-3 &1e-3 \\ \hline
\bf{SICI} &CP/QRA &Norm/CQN/OCQN &Jsu/CQJ/OCQJ &Stu/CQS/OCQS &QR/CQR/OCQR
\\ \hline
${n_h}$ &768 &768 &512 &960 &896 \\ 
${l_r}$ &1e-4 &1e-5 &1e-3 &1e-3 &1e-5 \\
\end{tabular}
\end{adjustbox}
\end{center}
\end{table}
The experiments have been performed by leveraging \texttt{Tensorflow}, including the \texttt{Tensorflow Probability} library which provides several utilities to implement distributional NNs.\footnote{{To setup the backbone DE framework, the description of the configurations follows the one used in the analysis reported in \cite{MARCJASZ2023106843} for the baseline distributional and quantile regression based methods.}}

In particular, we adopt a set of 4 DNNs including 2 hidden layers with softplus activations. Training is performed by means of Adam \cite{kingma2017adam} with early-stopping, starting from different random initializations. The number of units in each layer and the learning rate are tuned by cross-validation.
The CP calibration subset involve 182 samples (i.e., the preceding 6 months) as in the QRA benchmark to achieve comparable results.  

In principle, a wide range of alternative configurations could be considered during the setup of the learning framework and the hyper-parameters experiments. 
Since our goal is not to obtain the best NNs architecture but to evaluate the introduction of the conformal inference framework under coherent backbone DE setups, we performed a restricted grid search based cross-validation procedure. 
The following arrangements have been adopted\footnote{{Dropout and $\ell1/\ell2$ regularizations have not been included in the layers since almost never chosen by the hyper-parameter tuner in \cite{MARCJASZ2023106843}.}}. The batch size has been set to 64.
As the different output layer's parameterizations and loss functions may require specific complexities and rates, the hidden unit size and the learning tuning are searched in the discrete sets $n_h\in$ [64, 128, 512, 640, 768, 896, 960] and $l_r\in$ [1e-5, 1e-4, 1e-3, 1e-2] respectively. 
Clearly the selections equal in the subsequent conformal inference stages (as e.g. in Jsu/CQJ/OCQJ).
NNs training is performed by a maximum number of 800 epochs, including an early stopping callback on the validation loss with a patience of 50 epochs. For the hyperparameters of the OCQ* methods we have investigated a common set across the different datasets and coverage levels, defined as $K_I$=10, $\eta$=1e-2, $C_{sat}$=1.2 given $T$=1e9 and a burn-in of 7 steps. 

During the daily recalibration of out-of-sample test experiments, the oldest sample in the moving window is discarded, while leaving a 20\% subset to evaluate the loss for early stopping.
The output quantiles of the distributional NNs are estimated by generating 10000 samples for each test prediction. 

Table~\ref{hyper_tune} reports the hidden units and learning rate employed for the test set experiments, as selected by the grid search procedure for each bidding zones. 
Large $n_h$ have been chosen in most cases.
We did not observe sensible differences in validation losses among closed configurations (e.g., 640 vs 896). Still, the specific selections may be influenced by local minimizers reached by the training algorithm. These are marginalized by the Deep Ensemble during testing.\footnote{{Prediction with different model ensembling techniques is left for future extensions.}} 

\subsection{Evaluation metrics and tests}
To evaluate the probabilistic forecasting performances achieved on the test sets, we follow 
%the approach employed in \cite{MARCJASZ2023106843}, which conforms to 
the common practices in the PEPF field (see e.g. \cite{NOWOTARSKI20181548} and references therein for further details). Calibration is first analyzed by means of the Kupiec test (with significance level 0.05) for unconditional coverage, on both extreme and central PIs, for each day-ahead hour. Then, the average Pinball loss across the deciles (see section \ref{Methods}) and the Winkler’s score for the related miscoverage levels are employed as proper scoring rules.
It is worth noting that the average Pinball score, computed over the distribution percentiles, is commonly employed in PEPF as a cheap discrete approximation to the continuous ranked probability score (CRPS) (see e.g. \cite{NOWOTARSKI20181548})
Statistical significance is evaluated via the multivariate Diebold and Mariano (DM) test on the differences in the loss norms between competing models. Point prediction errors are evaluated beyond the probabilistic forecasting performances by means of the Mean Absolute Error (MAE):
\begin{equation}
    \text{MAE}=\frac{1}{N}\sum_{n=1}^{N} | \mathrm{y}_{n} - \hat{y}_{n} |
\end{equation}

\subsubsection{Kupiec test}
The Kupiec test is aimed to assess the unconditional coverage of the PIs to the nominal rate $1-\alpha$. Defining the sequence of indicators: $I_n={1}\{ \mathrm{y}_{n} \in [\hat{L}_n, \hat{U}_n] \}$, with $\hat{L}_n, \hat{U}_n$ lower and upper bounds respectively, it checks whether $\mathbb{P}(I_n=1)=1-\alpha$, assuming independent violations.

The test is performed on the likelihood ratio (LR) statistics for unconditional coverage, $\chi^2(1)$ asymptotically distributed, defined as:
\begin{equation}
\text{LR}_{\text{UC}}=\frac{\alpha^{n_0}(1-\alpha)^{n_1}}{(1-\pi)^{n_0}\pi^{n_1}} \text{,  with:  }\pi=n_1/(n_1+n_0) 
\end{equation}
where $n_1$,$n_0$ represent the number of hits and violations in the indicator series. 
We leave to previous reviews in the literature (see, e.g., \cite{WERON20141030}) for further details on the common exploitation of the Kupiec test within the PEPF field. 

\subsubsection{Winkler's score}
The Winkler's score (or interval score) is a proper scoring rule to assess probabilistic forecasts formed as Prediction Intervals (PI) at discrete coverage levels $1-\alpha$ (see \cite[section 5.9]{ha2021}).  
Formally, it is defined as:
\begin{equation}
\label{winkler}
    Winkler_n =\begin{cases}
        \delta_n, & \text{if:  }\mathrm{y}_{n} \in [\hat{L}_n, \hat{U}_n]\\
        \delta_n + \frac{2}{\alpha}(\hat{L}_n-\mathrm{y}_{n}), & \text{if:  }\mathrm{y}_{n} < \hat{L}_n\\
        \delta_n + \frac{2}{\alpha}(\mathrm{y}_{n}-\hat{U}_n), & \text{if:  }\mathrm{y}_{n} > \hat{U}_n\\
    \end{cases}
\end{equation}
where $\delta_n=\hat{U}_n-\hat{L}_n$ is the width the $1-\alpha$-PI. 
The first case in (\ref{winkler}) rewards narrow PIs (i.e., sharpness), while the others penalizes the occurrence of test observations outside the predicted interval.

\subsubsection{Diebold–Mariano (DM) test}
\label{DM_test}
The model-free Diebold–Mariano (DM) test is widely adopted in the EPF literature to evaluate the statistical significance of the performance differences between the model's predictions (see e.g., \cite{TSCHORA2022118752}, \cite{WERON20141030} and references therein for further details). 

Rather than averaging the prediction scores over the dataset - as in conventional test set metrics analysis - the DM test is based on the computation of pairwise score differentials between the forecasts $\hat{y}_{d}^{Mi}$ and $\hat{y}_{d}^{Mj}$ provided by competing models $M_i$ and $M_j$ over the test days $d=[1,...,N_d]$, followed by an asymptotic z-test to assess the null hypothesis that the expected value of the differentials series is zero. 

Two versions of the DM test exist, namely the univariate (executed hour-wise), and multivariate (performed jointly across all day-ahead hours). Following the aforementioned research studies, we exploit the latter for the present work. Notably, besides being more fitted to multi-hour prediction frameworks and related cross-hour combinations in the overall training loss functions, it enables a more user-friendly summary representation of the test results by graphical heat maps. 

Formally, the forecasts score differentials are computed as follows:
\begin{align}
	\begin{aligned}
		\Delta_d^{M_i,M_j} &= \mathcal{L}\left[   \varepsilon_d^{M_i} \right] -  \mathcal{L}\left[ \varepsilon_d^{M_j} \right] \ , \ \text{with: } \mathcal{L} \left[ \varepsilon_d^{M_i} \right]=\left( \sum_{h=1}^{24} | \varepsilon_{d,h}^{M_i} |^n \right)^{1/n} 
	\end{aligned}
\end{align}
where $\varepsilon_{d,h}^{M_i} = \left({y}_{d,h} - \hat{y}_{d,h}^{M_i} \right) \in \mathbb{R}$, and the arbitrary loss function $\mathcal{L}$ is stated in common $n$-norm form. 
Then, the DM statistic is derived as:
\begin{equation}
	\text{DM}^{M_i,M_j}=\sqrt{N_d}\frac{\hat{\mu}^{M_i,M_j}}{\hat{\sigma}^{M_i,M_j}}
\end{equation}
with $\hat{\mu}^{M_i,M_j}$ and $\hat{\sigma}^{M_i,M_j}$ depicting the sample mean and standard deviation of the differentials series $\Delta_d^{M_i,M_j}$ computed over an out-of-sample test period of length $N_d$. 
In practice, the $p$-values of two one-sided tests are often investigated, framed on the null hypothesis $\text{H}_0 : \mathbb{E}\left[\Delta_d^{M_i,M_j} \right] \leq 0$ and alternative $\text{H}_1 : \mathbb{E}\left[\Delta_d^{M_i,M_j} \right] > 0$.
Hence, the $\text{H}_0$ rejection suggests a statistically significant performance improvement in the predictions provided by model $M_i$ with reference to the one of model $M_j$. To this end, a $p$-value of 5$\%$ is conventionally adopted as the test threshold.

\subsection{Results analysis}
\begin{figure}[t!]
\begin{center}
%\framebox[4.0in]{$\;$}
\includegraphics[width=1.0\linewidth]{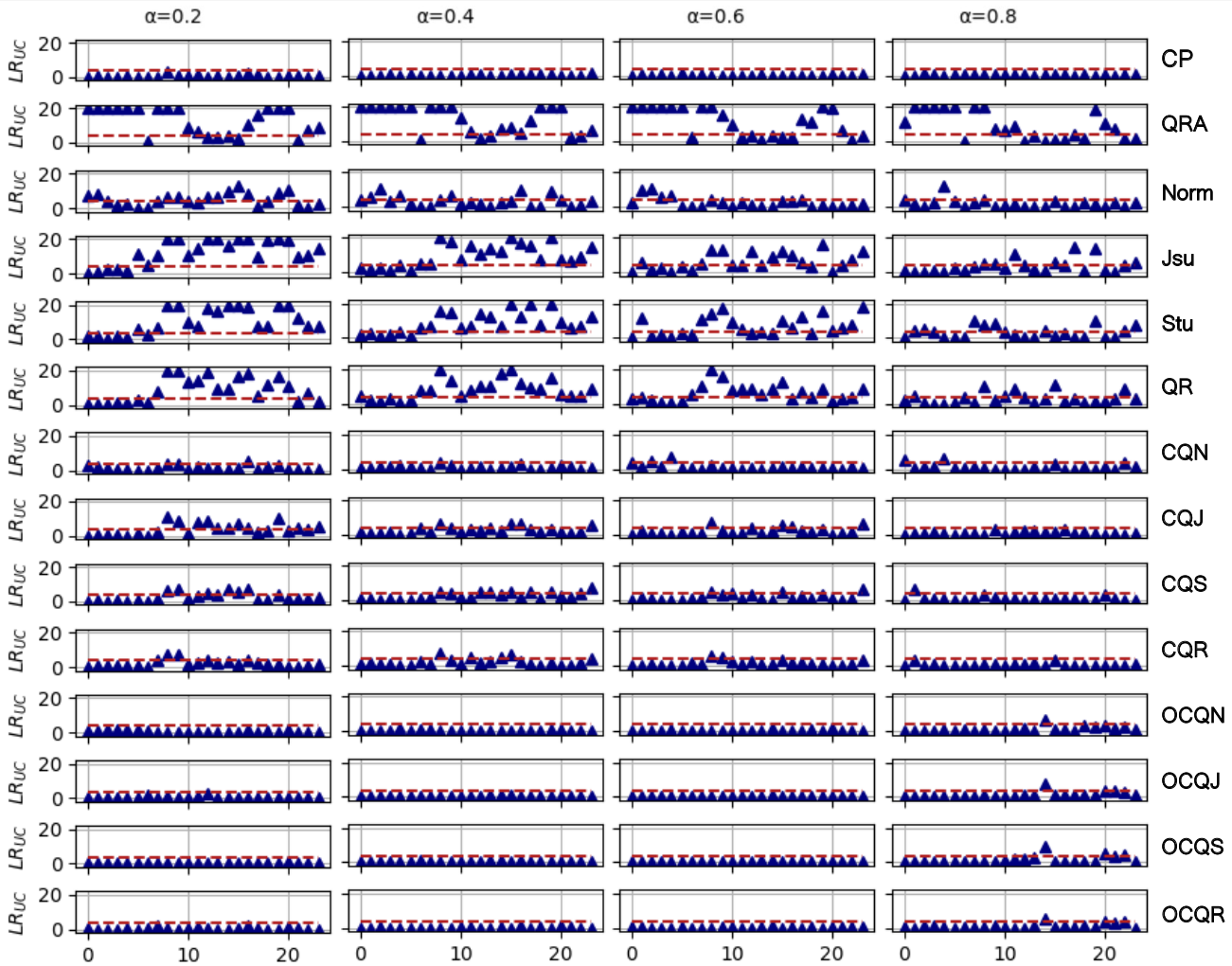}
\end{center}
\caption{Hourly Kupiec test on GE market test set}
\label{kupiec_test_plots_GE}
\end{figure}
\begin{figure}[t!]
\label{kupiec_test_plots_NOR}
\begin{center}
%\framebox[4.0in]{$\;$}
\includegraphics[width=1.0\linewidth]{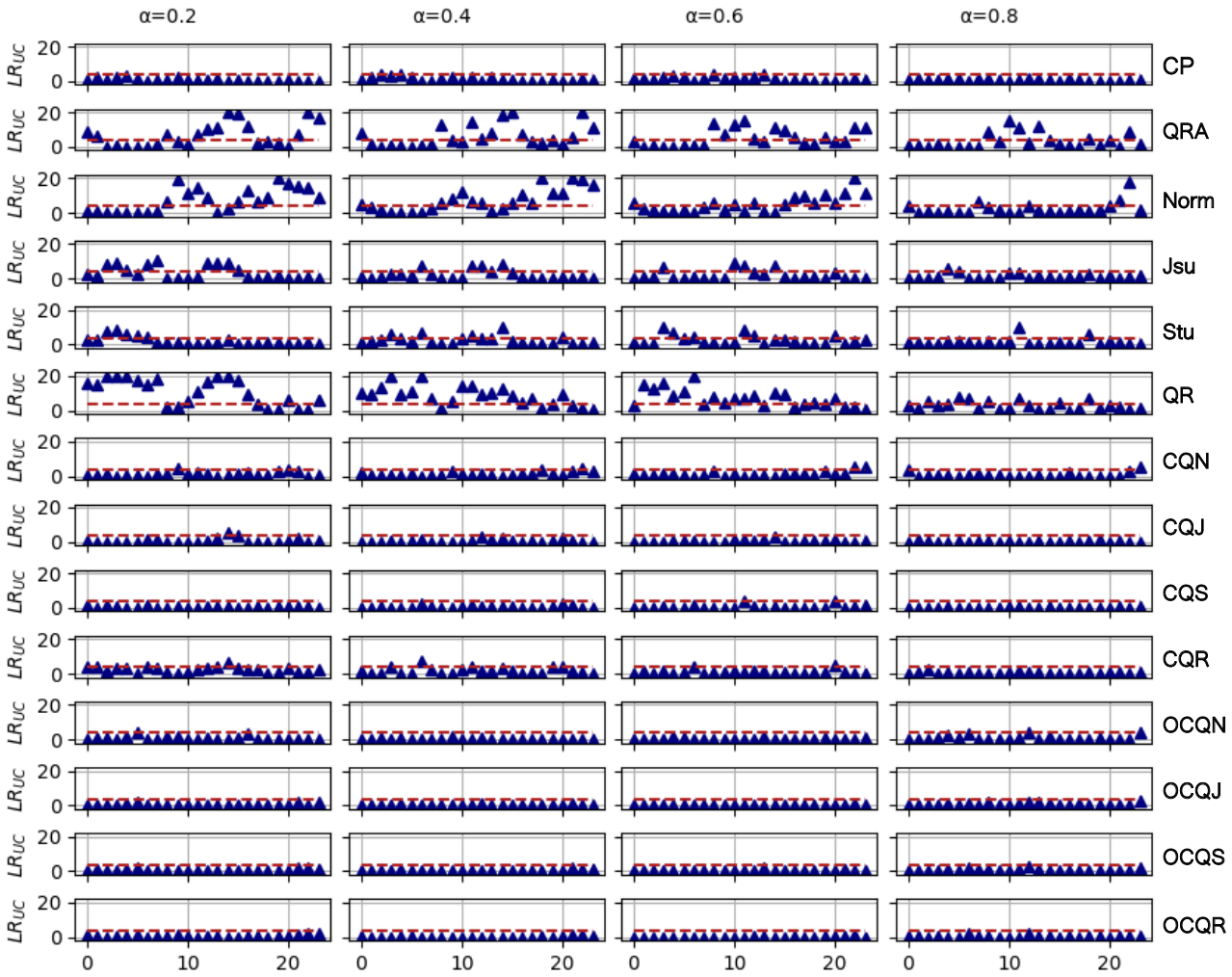}
\end{center}
\caption{Hourly Kupiec test on NOR market test set}
\end{figure}
\begin{figure}[t!]
\label{kupiec_test_plots_CNOR}
\begin{center}
%\framebox[4.0in]{$\;$}
\includegraphics[width=1.0\linewidth]{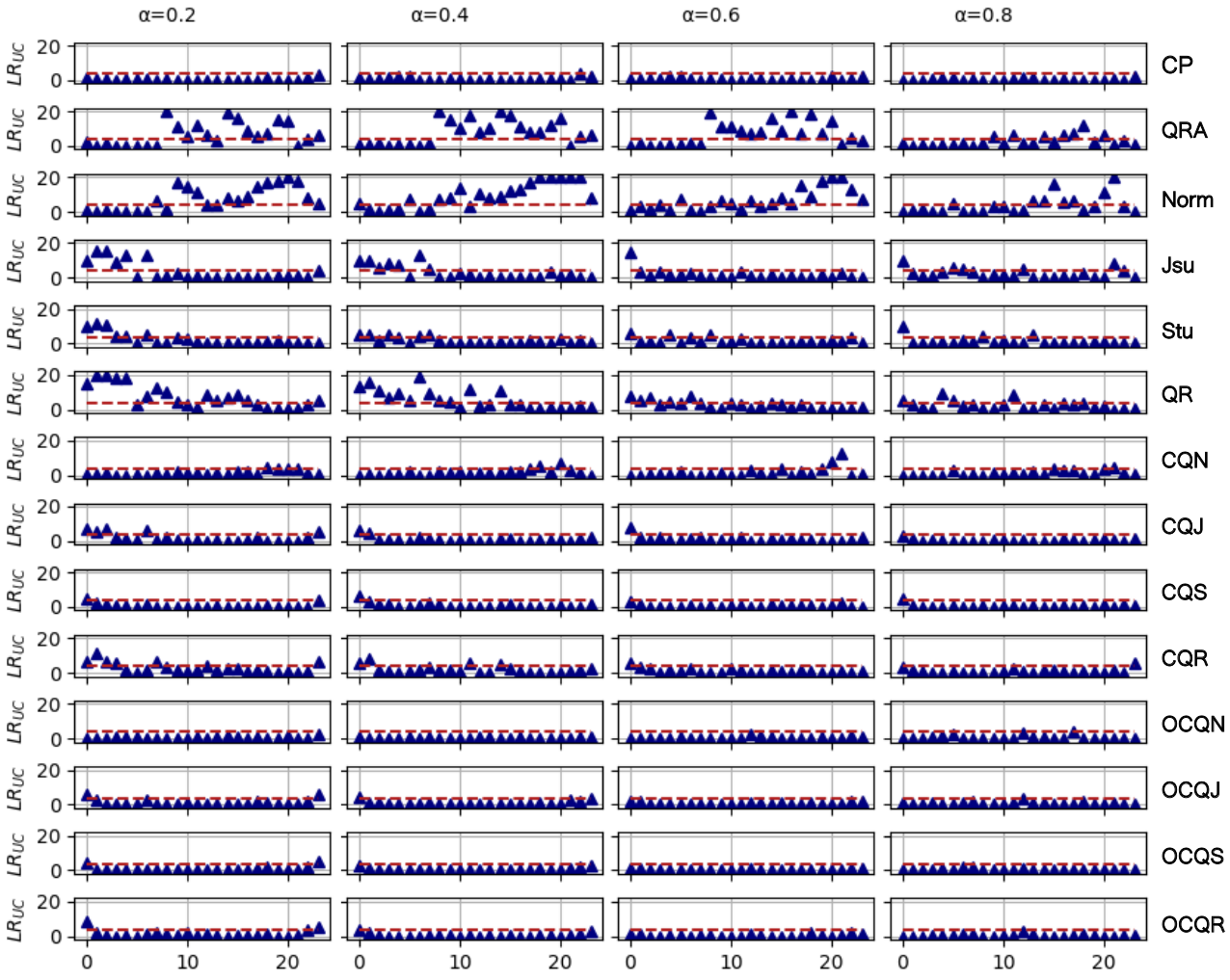}
\end{center}
\caption{Hourly Kupiec test on CNOR market test set}
\end{figure}
\begin{figure}[t!]
\label{kupiec_test_plots_CSUD}
\begin{center}
%\framebox[4.0in]{$\;$}
\includegraphics[width=1.0\linewidth]{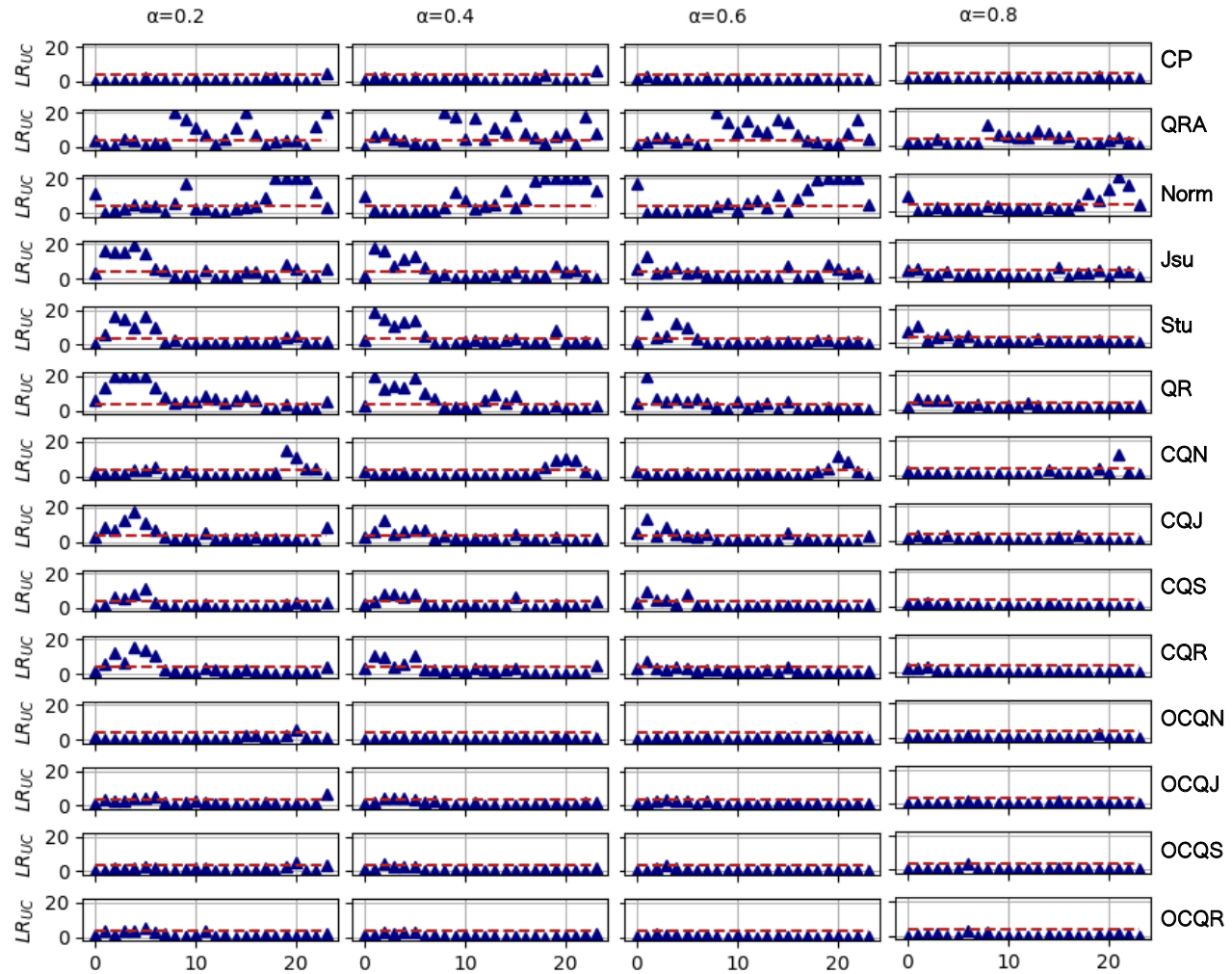}
\end{center}
\caption{Hourly Kupiec test on CSOU market test set}
\end{figure}
\begin{figure}[t!]
\label{kupiec_test_plots_SUD}
\begin{center}
%\framebox[4.0in]{$\;$}
\includegraphics[width=1.0\linewidth]{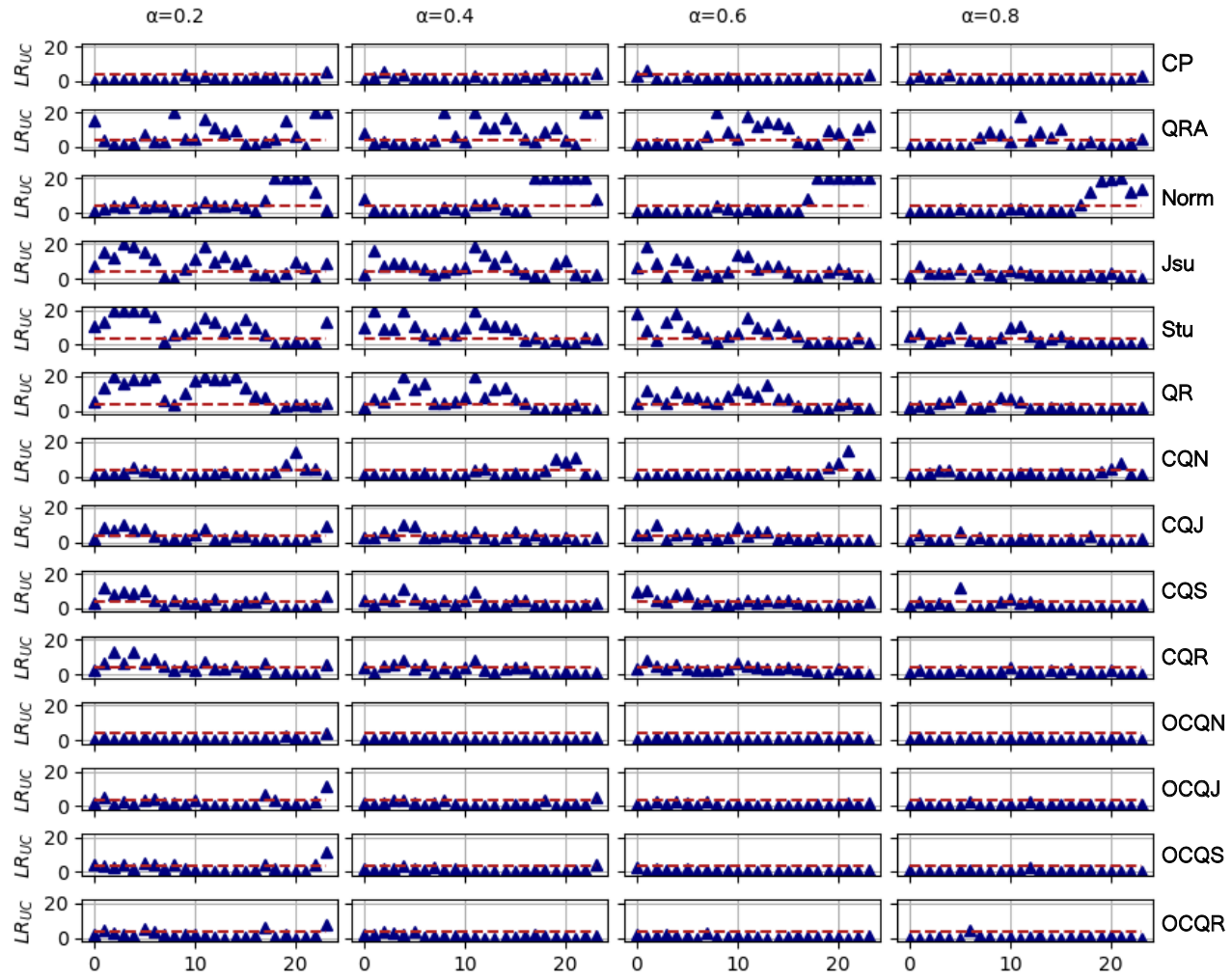}
\end{center}
\caption{Hourly Kupiec test on SOU market test set}
\end{figure}
\begin{figure}[t!]
\label{kupiec_test_plots_SARD}
\begin{center}
%\framebox[4.0in]{$\;$}
\includegraphics[width=1.0\linewidth]{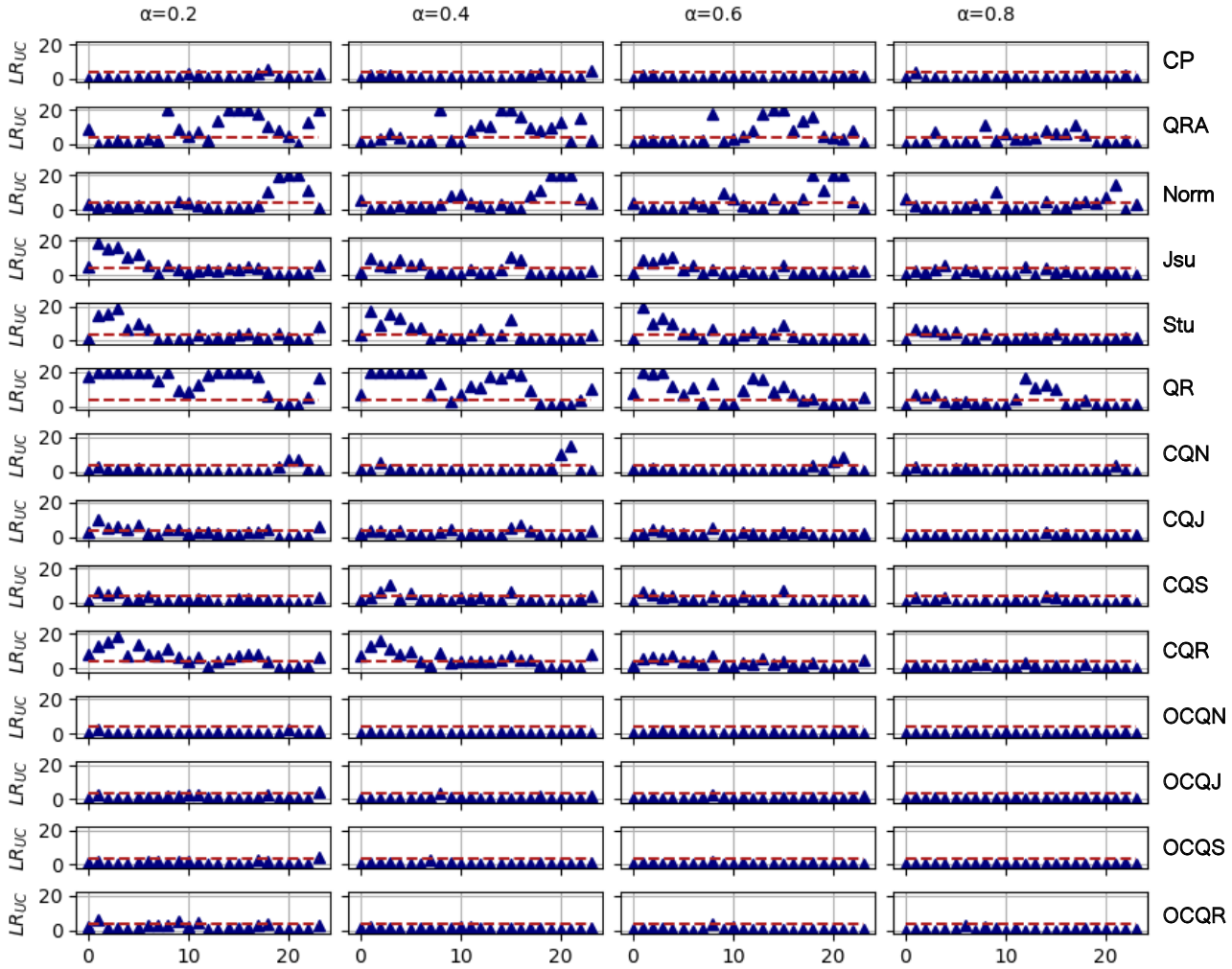}
\end{center}
\caption{Hourly Kupiec test on SARD market test set}
\end{figure}
\begin{figure}[t!]
\begin{center}
%\framebox[4.0in]{$\;$}
\includegraphics[width=1.0\linewidth]{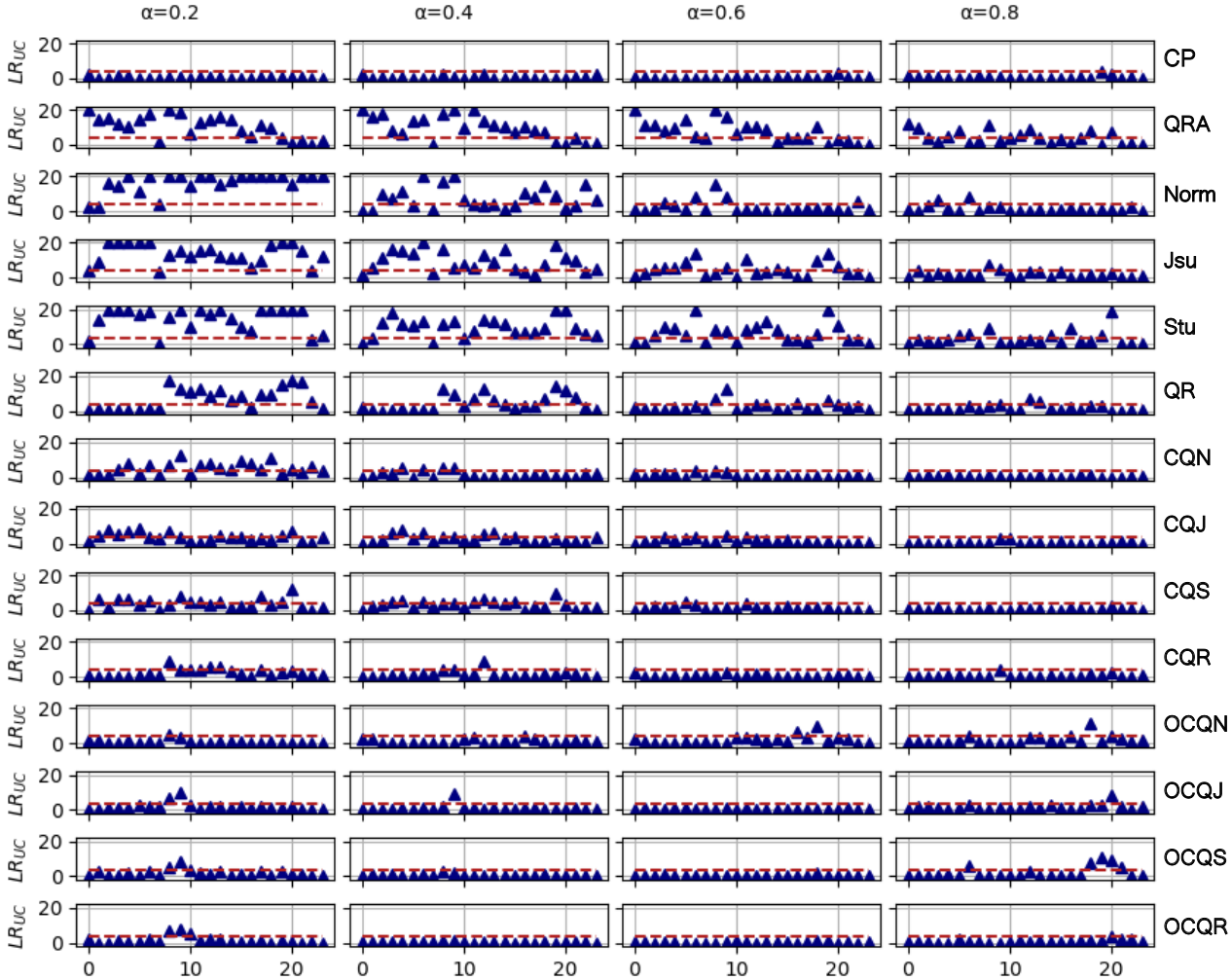}
\end{center}
\caption{Hourly Kupiec test on SICI market test set}
\label{kupiec_test_plots_SICI}
\end{figure}

The calibration tests of the benchmark methods, shown in Figure  \ref{kupiec_test_plots_GE}-\ref{kupiec_test_plots_SICI}, appear consistent with the results reported in previous studies. We observe worse coverage of Jsu and Stu on GE than other cases, which could be due to the reduced heterogeneity in the DE components impacting the quantification of forecast uncertainty \cite{BRUSAFERRI2022118341}.
It can be noted that the behaviour of the benchmarks differs between the regional markets, as shown e.g. in the Kupiec plots of NOR vs SICI. This could be related to task specific characteristics (e.g., volatility extent) and requires further investigations, e.g. by exploring alternative parameterized densities and datasets. 
The introduction of conformal inference has lead to improved hourly reliability on both distributional and QR settings. While the conventional CP on absolute residuals already provides adequate coverage, the effect of a more flexible approach is observed by assessing sharpness beyond calibration through the scoring rules (reported below). In fact, the former results in symmetric average bands around the model predictions, thus lacking the capability to adapt the PIs width on simpler/harder test conditions as required for sample-wise efficiency. 

\begin{figure}[t!]
\begin{center}
%\framebox[4.0in]{$\;$}
\includegraphics[width=1.0\linewidth]{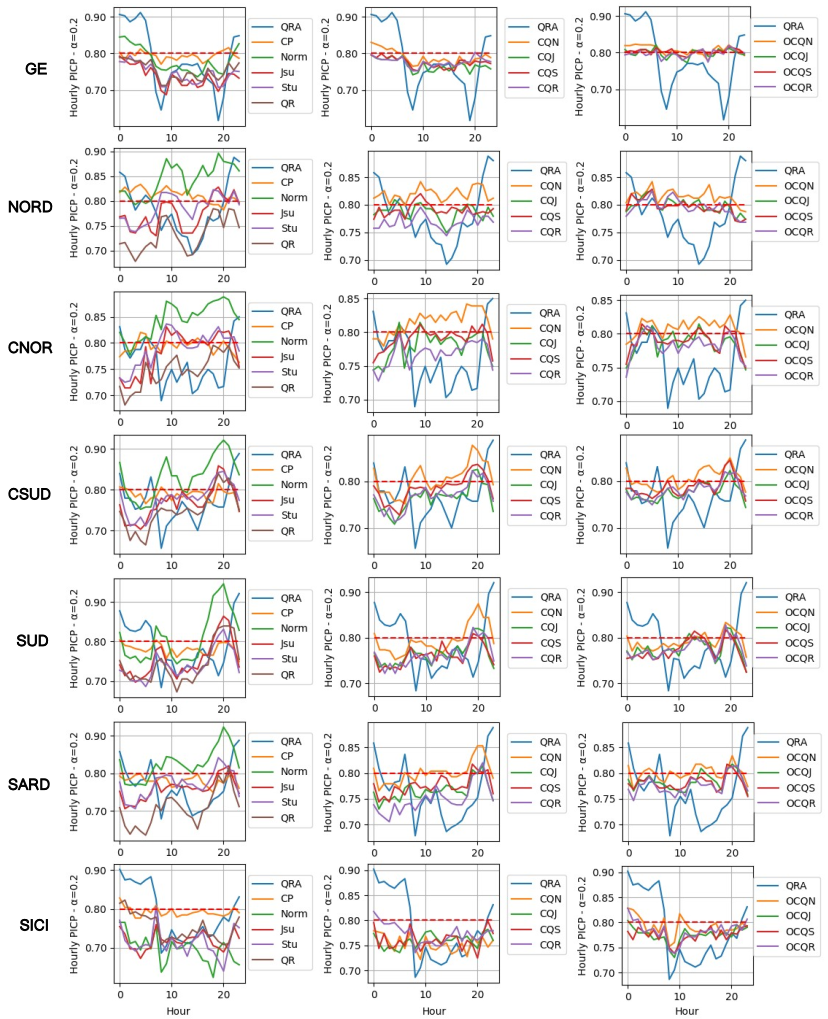}
\end{center}
\caption{Hourly PICP of miscoverege degree $\alpha=0.2$ on the test sets}
\label{hourly_PICP_plots_02}
\end{figure}

\begin{figure}[t!]
\begin{center}
%\framebox[4.0in]{$\;$}
\includegraphics[width=1.0\linewidth]{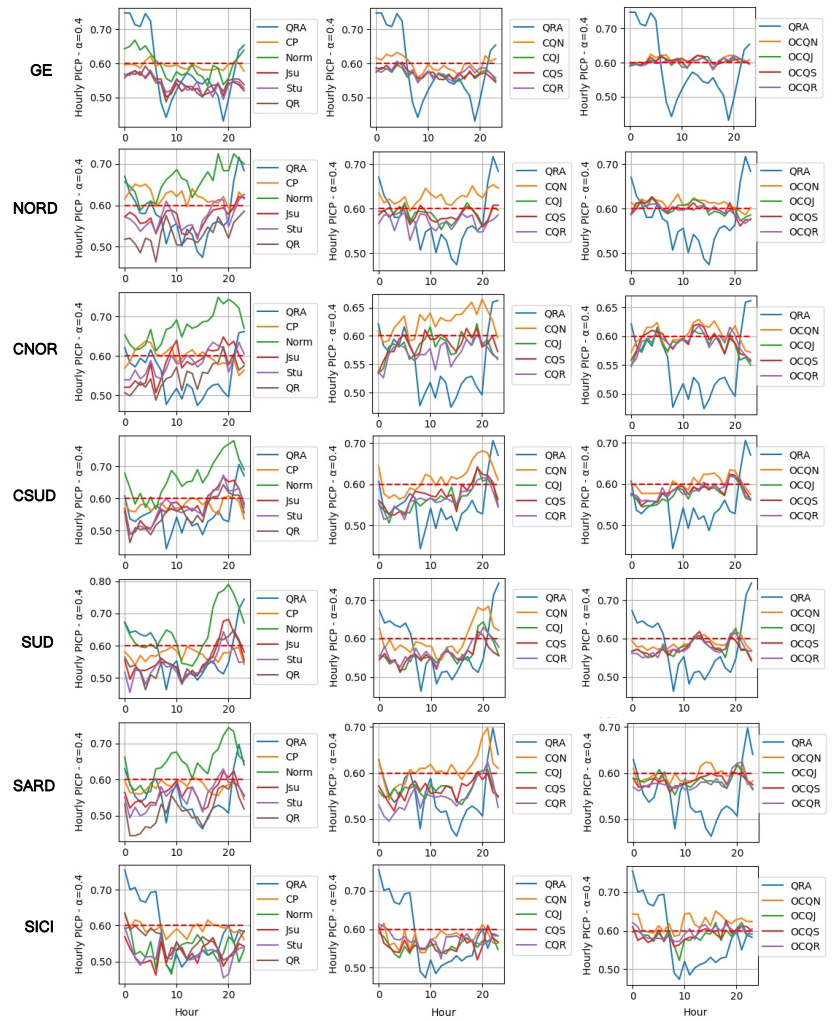}
\end{center}
\caption{Hourly PICP of miscoverege degree $\alpha=0.4$ on the test sets}
\label{hourly_PICP_plots_04}
\end{figure}

Further insights are provided by the hourly Prediction Interval Coverage Probability (PICP) values, reported in Figure~\ref{hourly_PICP_plots_02} and Figure \ref{hourly_PICP_plots_04} for $\alpha$=0.2 and $\alpha$=0.6 respectively, and the average PICPs for each coupled deciles in Table~\ref{PICP_table}. 
To support graphical readability, we have structured three subplots aggregating: the benchmark methods in the left column; the different conformalized quantile regression setups (CQ*) in the middle column; the on-line conformal PI control integration (OCQ*) in the right column. 
The QRA benchmark coverage is included in all subplots to provide a coherent base within the common y axis scale. 
The red dashed lines represent the target coverage value $1-\alpha$\footnote{e.g., $1-\alpha=0.8$ for $\alpha=0.2$}.


\begin{table}[t!]
\caption{Prediction Interval Coverage Probability computed for each 1-$\alpha$ coverage degree}
\label{PICP_table}
\begin{center}
\begin{adjustbox}{width=1\textwidth}
\small
\begin{tabular}{llllllllllllllll}
\toprule
& &      CP &     QRA &    Norm &     Jsu &     Stu &      QR &     CQN &     CQJ &     CQS &     CQR &    OCQN &    OCQJ &    OCQS &    OCQR \\
\midrule
   &PICP$_{\alpha=0.2}$          &   0.79 &   0.78 &   0.78 &   0.74 &   0.75 &   0.75 &   0.79 &   0.77 &   0.78 &   0.78 &   0.81 &   0.80 &   0.80 &   0.80 \\
GE &PICP$_{\alpha=0.4}$          &   0.59 &   0.59 &   0.60 &   0.54 &   0.54 &   0.54 &   0.60 &   0.57 &   0.57 &   0.57 &   0.61 &   0.60 &   0.61 &   0.60 \\
   &PICP$_{\alpha=0.6}$          &   0.40 &   0.39 &   0.40 &   0.35 &   0.35 &   0.35 &   0.40 &   0.38 &   0.38 &   0.38 &   0.41 &   0.40 &   0.40 &   0.40 \\
   &PICP$_{\alpha=0.8}$          &   0.20 &   0.20 &   0.20 &   0.17 &   0.18 &   0.18 &   0.20 &   0.19 &   0.19 &   0.19 &   0.22 &   0.21 &   0.21 &   0.21 \\
   \midrule
     &PICP$_{\alpha=0.2}$          &   0.81 &   0.78 &   0.84 &   0.77 &   0.79 &   0.73 &   0.82 &   0.79 &   0.79 &   0.77 &   0.81 &   0.80 &   0.80 &   0.80 \\
NOR &PICP$_{\alpha=0.4}$          &   0.62 &   0.57 &   0.66 &   0.57 &   0.57 &   0.53 &   0.63 &   0.59 &   0.59 &   0.57 &   0.61 &   0.60 &   0.60 &   0.60 \\
     &PICP$_{\alpha=0.6}$          &   0.43 &   0.38 &   0.45 &   0.38 &   0.37 &   0.34 &   0.42 &   0.39 &   0.39 &   0.38 &   0.41 &   0.40 &   0.40 &   0.40 \\
     &PICP$_{\alpha=0.8}$          &   0.22 &   0.18 &   0.22 &   0.19 &   0.18 &   0.17 &   0.21 &   0.20 &   0.19 &   0.19 &   0.21 &   0.21 &   0.21 &   0.21 \\
      \midrule
     &PICP$_{\alpha=0.2}$          &   0.79 &   0.76 &   0.85 &   0.78 &   0.79 &   0.75 &   0.82 &   0.78 &   0.79 &   0.77 &   0.81 &   0.79 &   0.79 &   0.78 \\
CNOR &PICP$_{\alpha=0.4}$          &   0.60 &   0.55 &   0.67 &   0.58 &   0.58 &   0.55 &   0.63 &   0.58 &   0.59 &   0.57 &   0.60 &   0.59 &   0.59 &   0.59 \\
     &PICP$_{\alpha=0.6}$          &   0.41 &   0.35 &   0.47 &   0.38 &   0.38 &   0.36 &   0.42 &   0.38 &   0.39 &   0.38 &   0.40 &   0.39 &   0.39 &   0.39 \\
     &PICP$_{\alpha=0.8}$          &   0.20 &   0.18 &   0.23 &   0.19 &   0.19 &   0.18 &   0.21 &   0.19 &   0.19 &   0.19 &   0.21 &   0.20 &   0.20 &   0.20 \\
      \midrule
     &PICP$_{\alpha=0.2}$          &   0.79 &   0.77 &   0.83 &   0.77 &   0.78 &   0.75 &   0.81 &   0.76 &   0.78 &   0.77 &   0.81 &   0.78 &   0.79 &   0.78 \\
CSOU &PICP$_{\alpha=0.4}$          &   0.58 &   0.55 &   0.67 &   0.58 &   0.57 &   0.55 &   0.61 &   0.57 &   0.57 &   0.56 &   0.60 &   0.58 &   0.58 &   0.58 \\
     &PICP$_{\alpha=0.6}$          &   0.39 &   0.36 &   0.46 &   0.38 &   0.38 &   0.37 &   0.42 &   0.37 &   0.37 &   0.37 &   0.40 &   0.38 &   0.39 &   0.39 \\
     &PICP$_{\alpha=0.8}$          &   0.20 &   0.18 &   0.24 &   0.19 &   0.19 &   0.18 &   0.21 &   0.19 &   0.19 &   0.19 &   0.21 &   0.20 &   0.20 &   0.20 \\
      \midrule
     &PICP$_{\alpha=0.2}$          &   0.78 &   0.79 &   0.81 &   0.75 &   0.74 &   0.74 &   0.80 &   0.76 &   0.76 &   0.76 &   0.79 &   0.78 &   0.77 &   0.78 \\
SOU  &PICP$_{\alpha=0.4}$          &   0.57 &   0.58 &   0.64 &   0.56 &   0.54 &   0.55 &   0.60 &   0.56 &   0.56 &   0.57 &   0.59 &   0.57 &   0.58 &   0.58 \\
     &PICP$_{\alpha=0.6}$          &   0.38 &   0.37 &   0.44 &   0.36 &   0.34 &   0.36 &   0.40 &   0.36 &   0.35 &   0.36 &   0.39 &   0.38 &   0.38 &   0.38 \\
     &PICP$_{\alpha=0.8}$          &   0.19 &   0.18 &   0.22 &   0.18 &   0.17 &   0.18 &   0.20 &   0.18 &   0.18 &   0.18 &   0.21 &   0.21 &   0.20 &   0.20 \\
      \midrule
     &PICP$_{\alpha=0.2}$          &   0.78 &   0.76 &   0.83 &   0.76 &   0.77 &   0.71 &   0.80 &   0.77 &   0.78 &   0.75 &   0.80 &   0.78 &   0.78 &   0.78 \\
SARD &PICP$_{\alpha=0.4}$          &   0.58 &   0.55 &   0.64 &   0.56 &   0.56 &   0.51 &   0.61 &   0.56 &   0.56 &   0.55 &   0.60 &   0.58 &   0.58 &   0.58 \\
     &PICP$_{\alpha=0.6}$          &   0.38 &   0.35 &   0.45 &   0.37 &   0.36 &   0.33 &   0.41 &   0.37 &   0.37 &   0.36 &   0.40 &   0.39 &   0.39 &   0.38 \\
     &PICP$_{\alpha=0.8}$          &   0.19 &   0.17 &   0.23 &   0.18 &   0.18 &   0.17 &   0.20 &   0.19 &   0.18 &   0.19 &   0.20 &   0.20 &   0.20 &   0.20 \\
      \midrule
     &PICP$_{\alpha=0.2}$          &   0.79 &   0.79 &   0.70 &   0.72 &   0.71 &   0.75 &   0.75 &   0.76 &   0.76 &   0.77 &   0.79 &   0.78 &   0.77 &   0.78 \\
SICI &PICP$_{\alpha=0.4}$          &   0.59 &   0.58 &   0.54 &   0.52 &   0.52 &   0.56 &   0.58 &   0.56 &   0.56 &   0.57 &   0.62 &   0.59 &   0.59 &   0.60 \\
     &PICP$_{\alpha=0.6}$          &   0.40 &   0.39 &   0.37 &   0.35 &   0.34 &   0.37 &   0.39 &   0.37 &   0.37 &   0.39 &   0.43 &   0.40 &   0.40 &   0.40 \\
     &PICP$_{\alpha=0.8}$          &   0.20 &   0.20 &   0.19 &   0.18 &   0.17 &   0.19 &   0.20 &   0.20 &   0.19 &   0.20 &   0.23 &   0.22 &   0.22 &   0.22 \\
\bottomrule
\end{tabular}
\end{adjustbox}
\end{center}
\end{table}

The baseline Jsu and Stu have obtained hourly coverage below the targets on average, which could be related to different uncertainties quantified on the samples observed during NNs training. Conversely, the Norm distributional form has resulted under-confident in several cases (see e.g., NORD and CNOR in Figure \ref{hourly_PICP_plots_02}). The hourly coverage of the CP approach on absolute residual agrees with the results of the Kupiec test reported above. 
The QR benchmark exhibit excessively narrow PIs, that can be attributed to both underlying quantile over-fitting phenomena during learning and data/distribution shifts. The combination in the ensemble seems to provide a limited contribution in fixing such issues.
Despite the marginal coverage degree, the QRA application on the DE shows sensible prediction step-specific coverage fluctuations. This can be motivated by the averaging effect of the backbone quantile regression performed across the hours. These observations worth further investigations, which is planned for future works.   

The introduction of the conformalized quantiles techniques has lead to improved PICPs throughout the different DE configurations. The comparison of the related columns in Figure~\ref{hourly_PICP_plots_02}-\ref{hourly_PICP_plots_04}, and Table~\ref{PICP_table} display the specific contributions towards the compensation of the limited calibration in the baseline models (see, e.g., QR w.r.t. CQR/OCQR in GE). 
Overall, the OCQ* settings show slightly better results than the CQ*. Still, we observe performance variations in OCQ* between different markets and coverage degrees (see, e.g., GE vs SICI). 
A possible explanation for this behavior could lie in the specific tuning of the proportional/integral actions hyper-parameters (see section \ref{non-exc}).  For the purpose of this work, we have employed simple common settings. Finer regulations (e.g., for each target coverage degree) may lead to more stable results across the testing conditions\footnote{This constitute a further interesting direction for future research.}. 

\begin{figure}[t!]
\begin{center}
%\framebox[4.0in]{$\;$}
\includegraphics[width=1.0\linewidth]{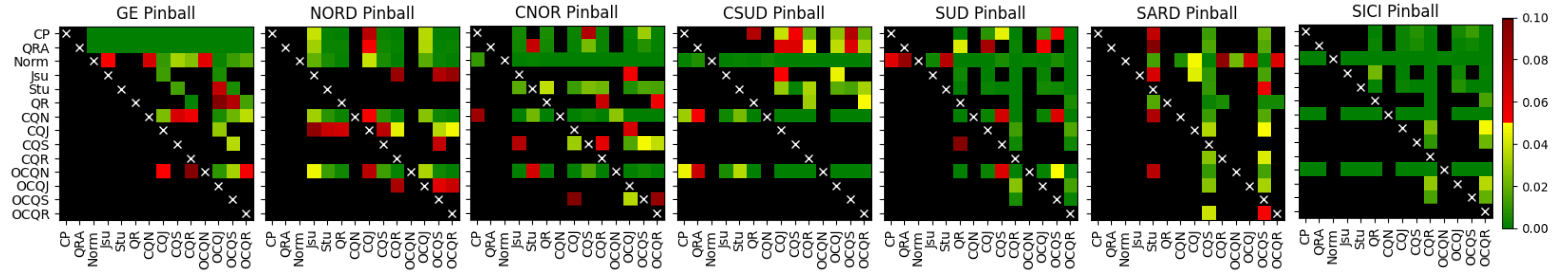}
\end{center}
\caption{DM test on average Pinball scores}
\label{DM_Pinball_fig}
\begin{center}
%\framebox[4.0in]{$\;$}
\includegraphics[width=1.0\linewidth]{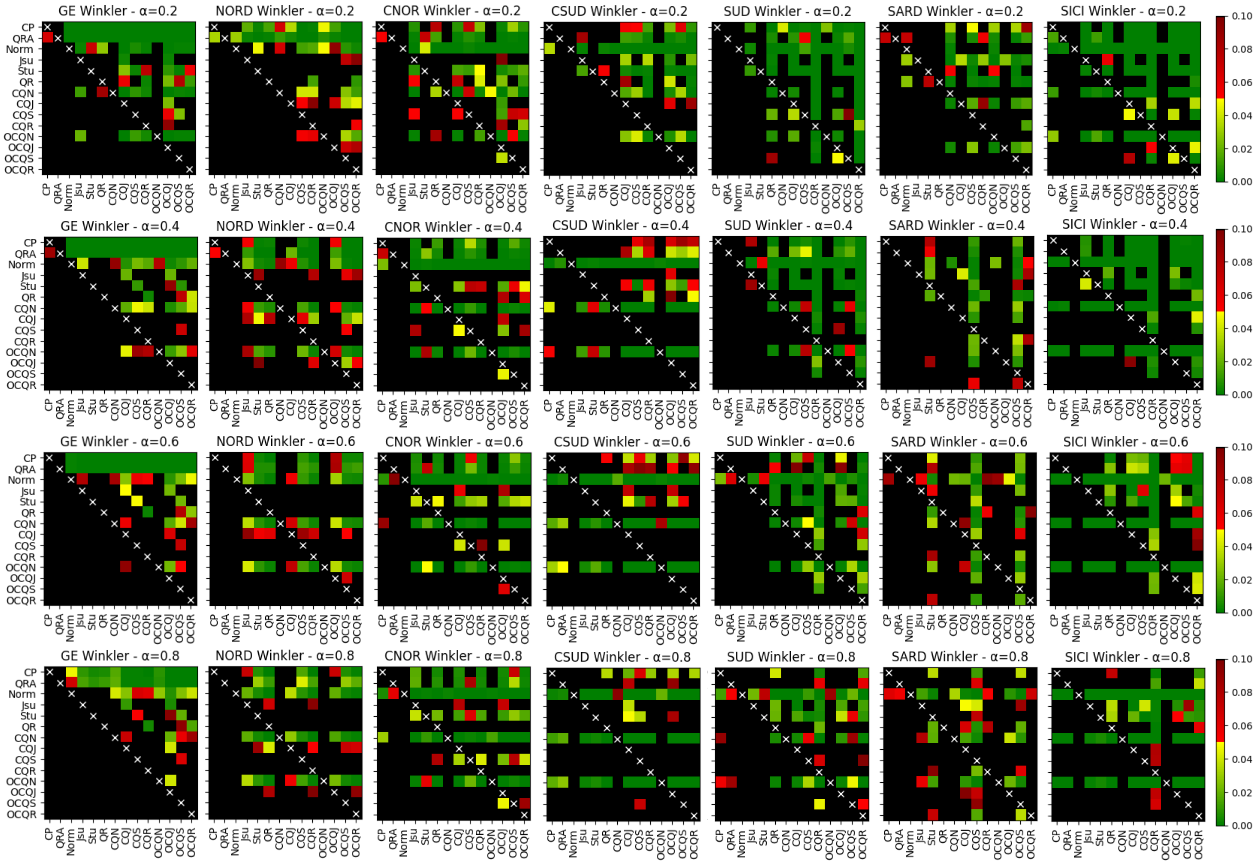}
\end{center}
\caption{DM test on Winkler's scores}
\label{Winklers_DM_test}
\begin{center}
%\framebox[4.0in]{$\;$}
\includegraphics[width=1.0\linewidth]{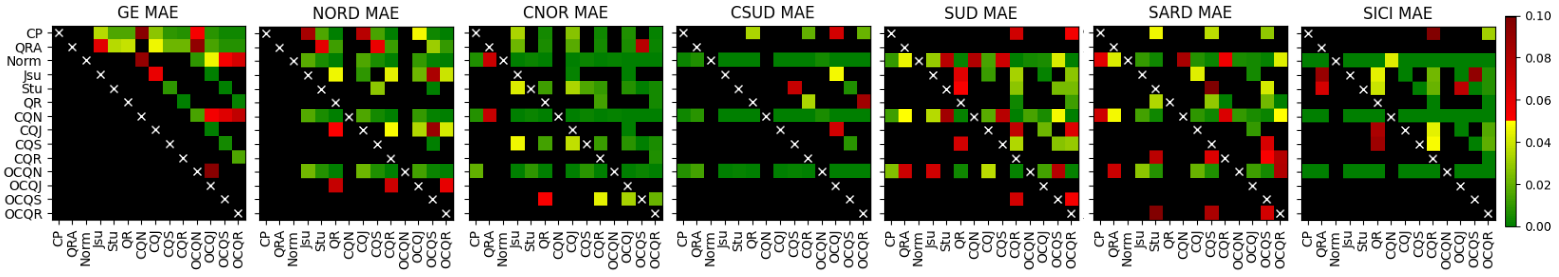}
\end{center}
\caption{DM test on MAE}
\label{DM_MAE}
\end{figure}

Beside being more reliable, the conformalized models have preserved stable and in some cases improved scores, as shown in Table~\ref{scores_table}.
Furthermore, the multivariate DM tests performed on the Pinball and Winkler's score are displayed in Figure~\ref{DM_Pinball_fig} and Figure~\ref{Winklers_DM_test} respectively, where the coloured cells highlight the p-value for the difference between model's predictions (see Section \ref{DM_test}). The test set Winkler's scores are computed for each of the PI$_{1-\alpha}$ obtained from the approximated distribution deciles.

\begin{table}[t!]
\caption{Test set scores for each market and PEPF technique}
\label{scores_table}
\begin{center}
\begin{adjustbox}{width=1\textwidth}
\small
\begin{tabular}{llllllllllllllll}
\toprule
& &      CP &     QRA &    Norm &     Jsu &     Stu &      QR &     CQN &     CQJ &     CQS &     CQR     &OCQN &OCQJ &OCQS &OCQR \\
\midrule
   &Pinball               &   1.549 &   1.557 &   1.489 &   1.467 &   1.473 &   1.474 &   1.483 &   1.457 &   1.463 &   1.462 &   1.480 &\bf{1.450} &   1.456 &   1.459 \\
   &Winkler$_{\alpha=0.2}$       &   20.27 &   20.62 &   18.07 &   17.52 &   17.77 &   17.71 &   17.99 &   17.30 &   17.53 &   17.50 &   17.98 &\bf{17.18} &   17.42 &   17.45 \\
GE &Winkler$_{\alpha=0.4}$       &   14.43 &   14.57 &   13.68 &   13.45 &   13.51 &   13.52 &   13.62 &   13.36 &   13.40 &   13.41 &   13.58 &   \bf{13.30} &   13.34 &   13.40 \\
   &Winkler$_{\alpha=0.6}$       &   11.39 &   11.45 &   11.08 &   10.94 &   10.97 &   11.00 &   11.03 &   10.88 &   10.91 &   10.91 &   11.02 & \bf{10.83} &   10.86 &   10.89 \\
   &Winkler$_{\alpha=0.8}$       &    9.27 &    9.28 &    9.13 &    9.04 &    9.05 &    9.06 &    9.09 &    8.97 &    9.00 &    9.00 &    9.06 &    \bf{8.92} &    8.96 &    8.97 \\
   &MAE                   &   3.799 &   3.794 &   3.758 &   3.724 &   3.725 &   3.726 &   3.755 &   3.720 &   3.718 &   3.717 &   3.745 &  \bf{3.700} &   3.704 &   3.704 \\
   \midrule
     &Pinball                &   1.847 &   1.861 &   1.855 &   1.823 &   1.809 &   1.806 &   1.852 &   1.827 &   1.810 &   1.804 &   1.850 &   1.822 &   1.804 &   \bf{1.803} \\
     &Winkler$_{\alpha=0.2}$       &   22.64 &   23.07 &   22.36 &   22.08 &   21.94 &   22.07 &   22.25 &   22.16 &   21.84 &   21.89 &   22.21 &   22.07 &   21.80 &\bf{21.79} \\
NOR &Winkler$_{\alpha=0.4}$       &   16.95 &   17.19 &   17.02 &   16.73 &   16.56 &   16.58 &   16.96 &   16.78 &   16.60 &   16.54 &   16.94 &   16.73 &\bf{16.52} &   16.53 \\
     &Winkler$_{\alpha=0.6}$       &   13.71 &   13.79 &   13.80 &   13.56 &   13.45 &   \bf{13.43} &   13.78 &   13.60 &   13.48 &   13.45 &   13.78 &   13.57 &   13.44 &   13.46 \\
     &Winkler$_{\alpha=0.8}$       &   11.31 &   11.34 &   11.41 &   11.19 &   11.11 &   \bf{11.06} &   11.41 &   11.22 &   11.13 &   11.08 &   11.39 &   11.19 &   11.10 &   11.08 \\
     &MAE                    &   4.655 &   4.659 &   4.705 &   4.609 &   4.581 &   4.547 &   4.705 &   4.608 &   4.580 &   4.547 &   4.700 &   4.601 &   4.565 &  \bf{4.543} \\
   \midrule
     &Pinball                &   1.963 &   1.979 &   2.007 &   1.925 &   1.952 &   1.927 &   1.989 &   1.920 &   1.943 &   1.922 &   1.981 &   \bf{1.913} &   1.935 &   1.918 \\
     &Winkler$_{\alpha=0.2}$       &   24.30 &   24.64 &   24.65 &   23.60 &   24.18 &   23.87 &   24.30 &   23.59 &   23.94 &   23.70 &   24.20 &   \bf{23.48} &   23.82 &   23.57 \\
CNOR &Winkler$_{\alpha=0.4}$       &   18.13 &   18.32 &   18.50 &   17.71 &   17.97 &   17.79 &   18.25 &   17.68 &   17.88 &   17.74 &   18.22 &   \bf{17.62} &   17.82 &   17.69 \\
     &Winkler$_{\alpha=0.6}$       &   14.56 &   14.66 &   14.90 &   14.29 &   14.46 &   14.29 &   14.75 &   14.23 &   14.39 &   14.26 &   14.71 &  \bf{14.20} &   14.34 &   14.25 \\
     &Winkler$_{\alpha=0.8}$       &   11.96 &   12.03 &   12.27 &   11.78 &   11.91 &   11.75 &   12.18 &   11.72 &   11.87 &   11.72 &   12.11 &   \bf{11.69} &   11.82 &   11.70 \\
     &MAE                   &   4.918 &   4.952 &   5.047 &   4.851 &   4.908 &   4.830 &   5.046 &   4.848 &   4.906 &   4.828 &   5.025 &   4.825 &   4.886 &  \bf{4.815} \\
   \midrule
     &Pinball               &   1.994 &   1.996 &   2.050 &   1.977 &   1.990 &   1.976 &   2.033 &   1.964 &   1.968 &   1.963 &   2.028 &   \bf{1.962} &   1.970 &   1.964 \\
     &Winkler$_{\alpha=0.2}$       &   24.88 &   25.07 &   25.42 &   24.60 &   25.04 &   24.66 &   24.92 &   24.36 &   24.37 &   24.14 &   24.88 &   24.24 &   24.36 &   \bf{24.09} \\
CSOU &Winkler$_{\alpha=0.4}$       &   18.38 &   18.45 &   18.96 &   18.27 &   18.40 &   18.29 &   18.73 &   18.17 &   18.13 &   18.15 &   18.71 &   \bf{18.12} &   18.14 &   18.14 \\
     &Winkler$_{\alpha=0.6}$       &   14.78 &   14.77 &   15.21 &   14.64 &   14.73 &   14.62 &   15.10 &   14.55 &   14.59 &   14.57 &   15.04 &   \bf{14.54} &   14.61 &   14.59 \\
     &Winkler$_{\alpha=0.8}$       &   12.14 &   12.12 &   12.47 &   12.03 &   12.07 &   12.00 &   12.42 &   {11.93} &   11.99 &   11.97 &   12.38 &   11.96 &   12.03 &   11.99 \\
     &MAE                   &   4.992 &   4.982 &   5.127 &   4.949 &   4.966 &   4.934 &   5.124 &   4.942 &   4.958 &  \bf{4.927} &   5.111 &   4.933 &   4.949 & \bf{4.927} \\
   \midrule
     &Pinball               &   2.176 &   2.170 &   2.209 &   2.164 &   2.178 &   2.130 &   2.184 &   2.138 &   2.152 &   2.110 &   2.185 &   2.136 &   2.150 &   \bf{2.109} \\
     &Winkler$_{\alpha=0.2}$       &   27.48 &   27.21 &   27.37 &   26.62 &   27.18 &   26.00 &   26.86 &   26.18 &   26.60 &   25.45 &   26.76 &   26.07 &   26.45 & \bf{25.26} \\
SOU  &Winkler$_{\alpha=0.4}$       &   20.18 &   20.12 &   20.45 &   19.93 &   20.12 &   19.60 &   20.09 &   19.66 &   19.79 &   19.35 &   20.09 &   19.60 &   19.76 & \bf{19.32} \\
     &Winkler$_{\alpha=0.6}$       &   16.09 &   16.07 &   16.39 &   16.07 &   16.13 &   15.80 &   16.19 &   15.85 &   15.93 & \bf{15.68} &   16.24 &   15.87 &   15.94 &   15.69 \\
     &Winkler$_{\alpha=0.8}$       &   13.17 &   13.16 &   13.44 &   13.22 &   13.24 &   13.05 &   13.33 &   13.06 &   13.12 &  \bf{12.98} &   13.37 &   13.07 &   13.14 &   12.99 \\
     &MAE                   &   5.414 &   5.407 &   5.528 &   5.440 &   5.447 &   5.384 &   5.525 &   5.426 &   5.444 &   5.373 &   5.511 &   5.409 &   5.431 &  \bf{5.368} \\
   \midrule
     &Pinball               &   2.258 &   2.261 &   2.282 &   2.262 &   2.234 &   2.275 &   2.266 &   2.252 &   \bf{2.218} &   2.255 &   2.267 &   2.254 &   2.221 &   2.251 \\
     &Winkler$_{\alpha=0.2}$       &   28.14 &   28.53 &   27.92 &   28.46 &   28.09 &   28.58 &   27.56 &   28.26 &   27.61 &   27.89 &   27.55 &   28.21 &  \bf{27.54} &   27.66 \\
SARD &Winkler$_{\alpha=0.4}$       &   20.84 &   20.93 &   21.01 &   20.97 &   20.59 &   21.00 &   20.82 &   20.86 &  \bf{20.42} &   20.79 &   20.83 &   20.88 &   20.45 &   20.71 \\
     &Winkler$_{\alpha=0.6}$       &   16.73 &   16.73 &   16.95 &   16.73 &   16.52 &   16.85 &   16.84 &   16.66 &  \bf{16.41} &   16.74 &   16.86 &   16.70 &   16.45 &   16.75 \\
     &Winkler$_{\alpha=0.8}$       &   13.75 &   13.70 &   13.96 &   13.71 &   13.57 &   13.81 &   13.88 &   13.63 &  \bf{13.51} &   13.74 &   13.90 &   13.67 &   13.55 &   13.76 \\
     &MAE                   &   5.653 &   5.624 &   5.745 &   5.632 &   5.584 &   5.672 &   5.741 &   5.627 &   5.580 &   5.654 &   5.727 &   5.617 &  \bf{5.575} &   5.647 \\
   \midrule
     &Pinball               &   4.433 &   4.440 &   4.646 &   4.396 &   4.428 &   4.341 &   4.581 &   4.346 &   4.362 & \bf{4.295} &   4.584 &   4.348 &   4.360 &   4.302 \\
     &Winkler$_{\alpha=0.2}$       &   56.78 &   57.86 &   60.44 &   54.65 &   56.10 &   53.78 &   58.37 &   53.63 &   54.33 & \bf{52.71} &   58.14 &   53.55 &   54.23 &   52.74 \\
SICI &Winkler$_{\alpha=0.4}$       &   41.45 &   41.58 &   43.35 &   40.66 &   41.13 &   40.09 &   42.85 &   40.11 &   40.39 & \bf{39.55} &   43.02 &   40.25 &   40.39 &   39.66 \\
     &Winkler$_{\alpha=0.6}$       &   32.75 &   32.70 &   34.22 &   32.61 &   32.74 &   32.23 &   33.83 &   32.28 &   32.30 &  \bf{31.91} &   33.91 &   32.33 &   32.36 &   32.01 \\
     &Winkler$_{\alpha=0.8}$       &   26.62 &   26.54 &   27.83 &   26.72 &   26.75 &   26.43 &   27.43 &   26.42 &   26.43 &  \bf{26.20} &   27.47 &   26.44 &   26.47 &   26.27 \\
     &MAE                   &  10.91 &  10.86 &  11.41 &  10.98 &  10.99 &  10.86 &  11.40 &  10.96 &  10.96 &  10.84 &  11.36 &  10.90 &  10.89 & \bf{10.81} \\
\bottomrule
\end{tabular}
\end{adjustbox}
\end{center}
\end{table}
Among the alternative configurations, the parameterized Normal form has obtained the worst performances, while CQR shows slightly better capabilities than CQJ and CQS. Similar results can be observed for the respective OCQ* settings. 
Such observations may be motivated by the major adaptivity, beyond the predetermined distribution forms, in characterizing complex uncertainty patterns following the specific applications needs. Besides, from the inspection of the Winkler's score, it appears that CP tends to contribute more to the correction of the PIs closed to the tails, which could be related to the behaviour of the semi-parametric quantiles estimation stage in regions with limited observations. This is visible e.g., in the DM plots of the Winkler's PI with $\alpha=0.2$ vs $\alpha=0.6$ regarding NOR and CSOU regions.
However, these are just initial explanations and worth further empirical analysis.

Besides the probabilistic forecasting performances, Table~\ref{scores_table} reports the point prediction Mean Absolute Errors, while the outputs of the related DM tests are depicted in Figure \ref{DM_MAE}. We observe that conformal inference has lead to a slight improvement of the MAE in some cases (see e.g. CNOR), as the median prediction can be updated after quantiles correction by the sorting procedure. The impact on point prediction accuracy of more flexible distribution parameterization have been reported also in \cite{MARCJASZ2023106843}. The gap in the MAE on GE could be related to the reduced input features set.

\begin{figure}[t!]
\begin{center}
%\framebox[4.0in]{$\;$}
\includegraphics[width=0.9\linewidth]{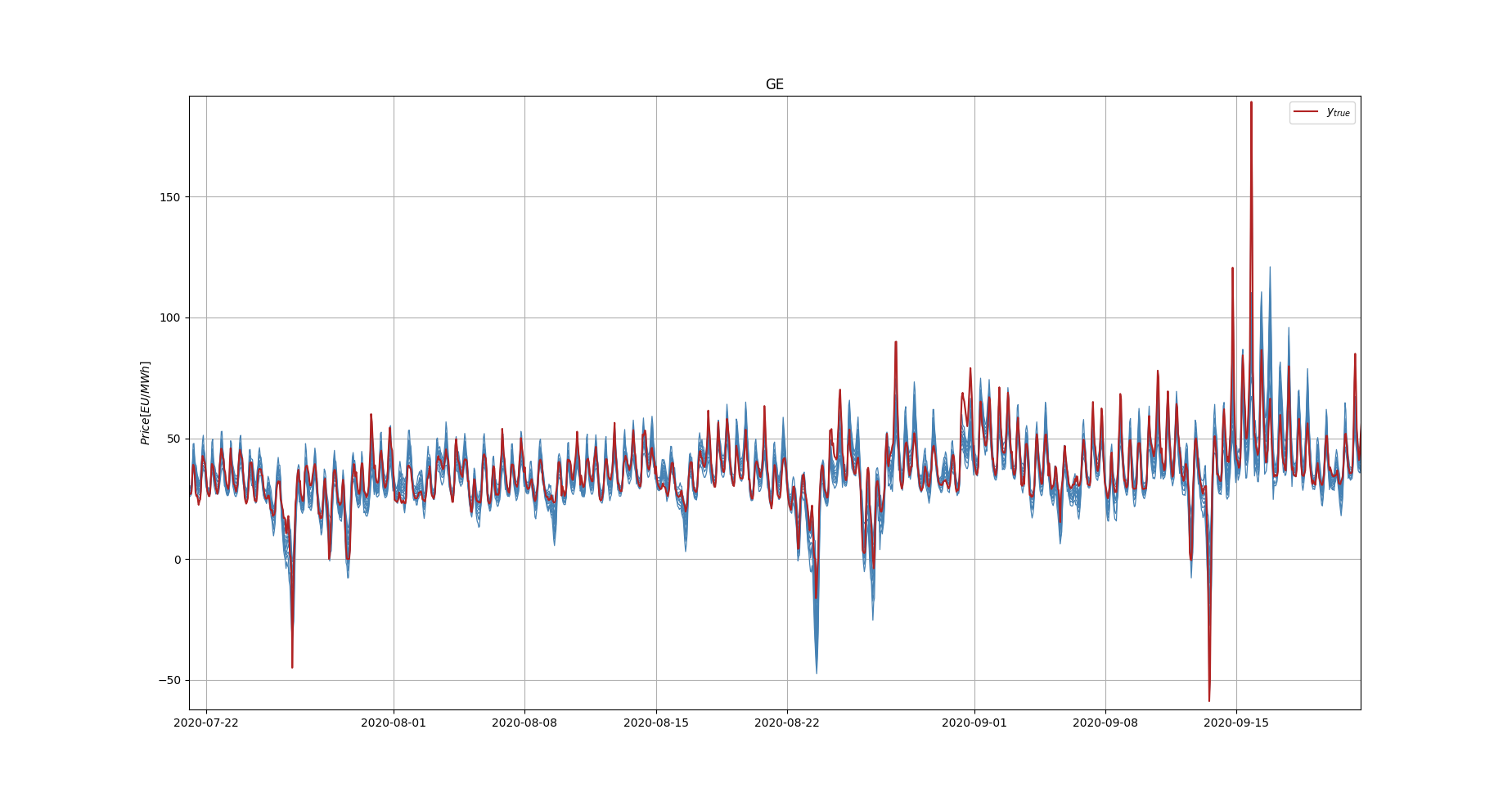}
\includegraphics[width=0.9\linewidth]{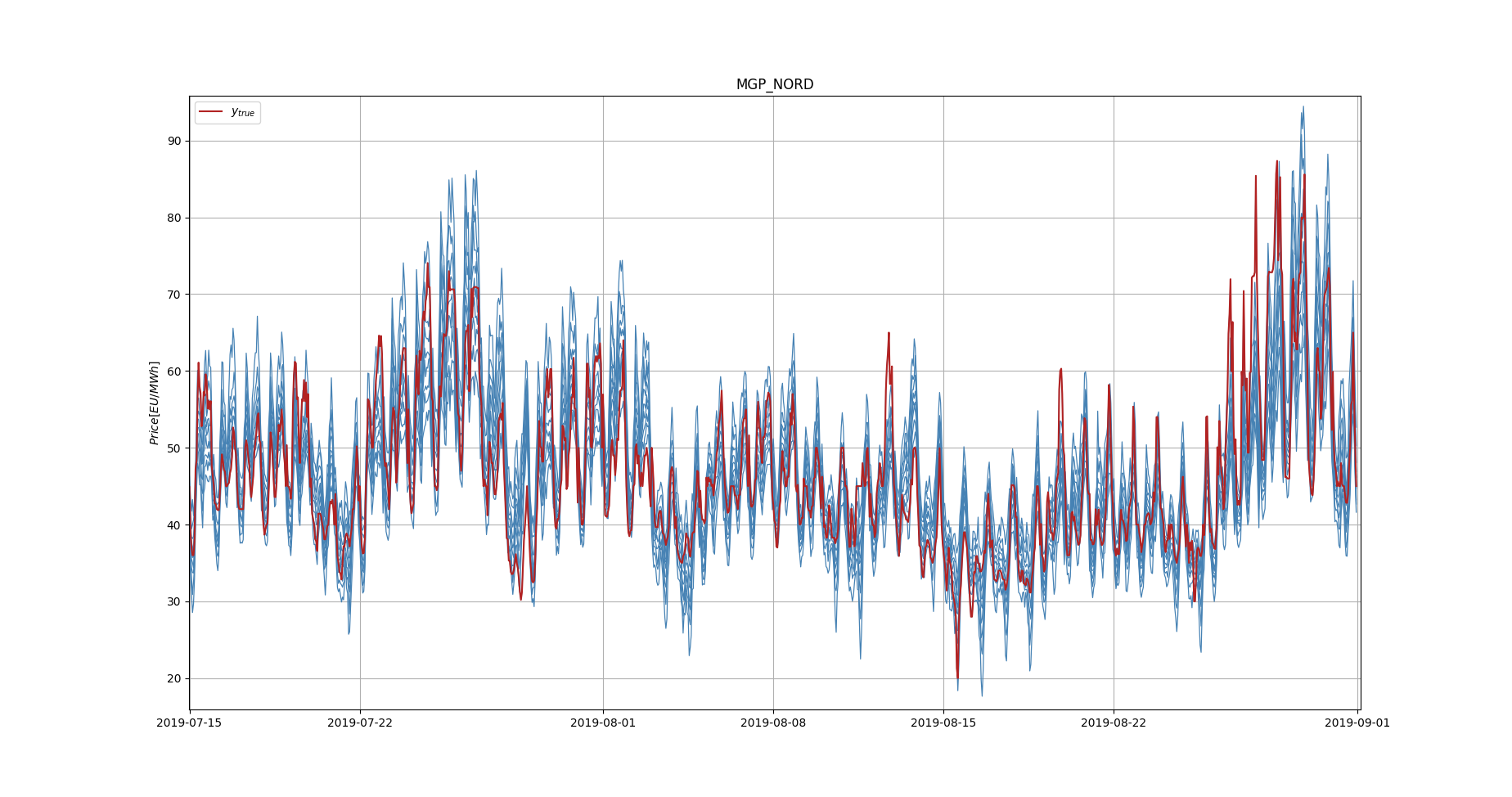}
\includegraphics[width=0.9\linewidth]{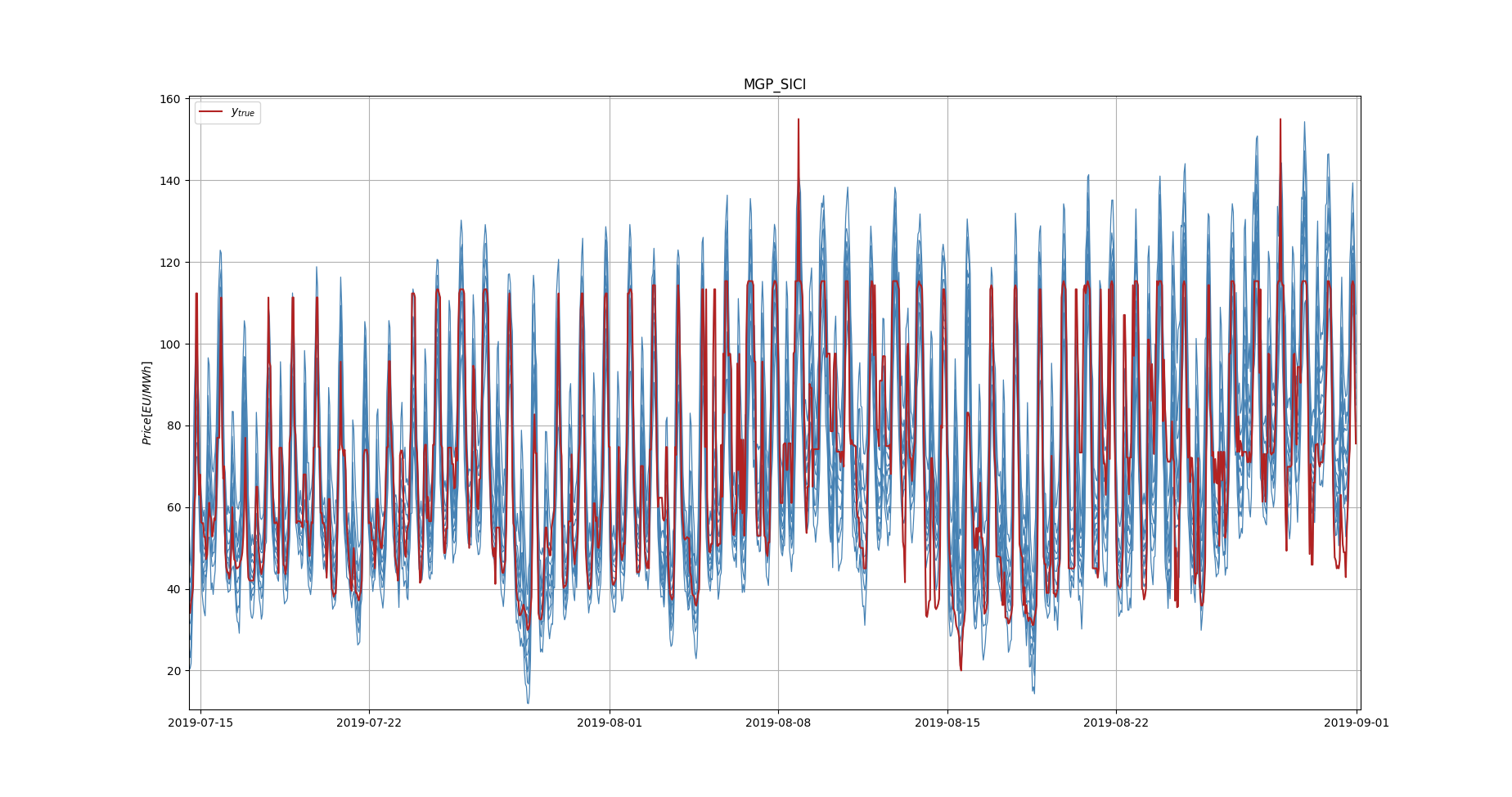}
\end{center}
\caption{Probabilistic forecasts on a subset of the test samples: (t)-GE, (m)-NOR, (b)-SICI}
\label{test_forec}
\end{figure}

To provide further insights on the different price behaviours addressed by the models among the regional markets, we plot in Figure \ref{test_forec} extracts from GE, NOR and SICI. The deciles predicted by the CQR model are reported besides the target series. Clearly, the price in each zone assume specific shapes and values, which depends on the generation mix (e.g., share of wind power generation in the southern part of Italy) and the different power demands. We leave to \cite{forecast5010003} and references therein for further details on the specific characteristics of the different Italian bidding area, and to \cite{MADADKHANI2024107241} for a recent detailed analysis of the German power market.  

	\section{Conclusions and next developments}
	The goal of PEPF, as probabilistic forecasting in general, is to maximize sharpness while respecting calibration requirements. 
In fact, efficient but reliabile uncertainty quantifications are decisive in supporting decision making under stochastic conditions, as it is the case in modern day-ahead EPF applications.
In this work, we addressed the limited hourly calibration of the state of the art neural network (NN) ensemble based methods for PEPF by exploiting a flexible Conformal Prediction (CP) framework.
To this end, the quantiles approximated by the NNs at each prediction step have been incrementally conformalized over the test set following a daily retraining (i.e., recalibration). Both quantile regression and distributional NNs approaches have been exploited for estimating the conditional quantiles to be calibrated.  
A uniform vincentization technique has been employed to aggregate the quantiles produced by the underlying ensemble components, providing a simple but robust forecasting combination in practice. Besides, a post-hoc sorting operator has been included to achieve conditional quantile non-crossing.
The asymmetric Conformalized Quantile Regression formulation, adjusted in a multi-horizon forecasting setup, have been deployed to compute the scores and the conformalized quantiles. 
This supports a proper correction of both upper and lower bounds at each stage over the prediciton horizon, leading to a more efficient local overage in the features conditioned PIs estimation. Such capability is particularly crucial to tackle the complex distributional patterns typically observed in PEPF applications, such as sensible heteroskedasticity, skewness and fat tails. Besides, it compensates potential quantile overfitting of the conventional pinball loss based neural network training, enhancing the asymptotic validity with finite samples coverage guarantees. To address the lack of robustness to failures of the samples exchangeability assumptions, we have exploited an adversarial setup. Quantiles tracking and coverage error integration stages have been included, aimed to compensate potential coverage degradation and systematic errors occurring over the recalibration window, e.g., due to sensible short-term drifts. 
The whole procedure have been started on a samples subset close to the first test date to acquire the initial bag of calibration scores, then proceeding in a rolling window fashion.
Indeed, updating the ensemble components by including the most recent observations support probabilistic performances beside point accuracy, as the underlying model quality impacts the local reliability of the prediction intervals despite the marginal coverage guarantees provided by the CP framework. 
Besides, the ensemble aggregation provides a further mechanisms for this purpose, by getting rid of poor local minimizers in the network parameter space reached by single trainings. 

The experiments have been executed on the German market dataset as well as on the different regional bidding zones constituting the Italian day-ahead markets, providing a compelling setup for testing under heterogeneous conditions. 
A comparison against state of the art benchmarks have been performed, including QRA, quantile regression NNs, distributional NNs, as well as conventional absolute score and normalized score based CP settings, showing the capability of the proposed approach to achieve day-ahead probabilistic forecasts with improved hourly coverage. Beside being more reliable, the conformalized models have been shown to preserve stable and in some cases improved probabilistic scores.

Still, we envision several avenues of future research, including the integration of further adaptive CP wrappers to hold coverage under data and concept drifts, mechanisms for fine-tuning of the coverage levels, CP on early-stopping for improving sample-efficiency, as well as further baseline models, ensembles architectures and datasets. The deployment of computationally cheaper deep ensemble approaches represents an important issue to be addressed, to reduce the resource consumption of multiple network training during recalibration. 
Furthermore, we plan to implement eXplainable AI techniques (e.g., SHAP) to investigate the major features affecting the probabilistic predictions.
Further to this, despite the specific energy forecasting tasks addressed in this study, the developed approach can also be extended to other applications, such as day-ahead electricity load forecasting.
		
	%\section*{References}
	\bibliography{mybibfile_v01}
	
\end{document}